\documentclass[journal,lettersize]{IEEEtran}
\usepackage{algorithmic}
\usepackage{algorithm}
\usepackage{amsmath,amsfonts,amsthm,amssymb}
\usepackage[colorlinks,
            linkcolor=black,
            anchorcolor=black,
            citecolor=black,
            urlcolor=black]{hyperref}
\usepackage{multirow}
\usepackage{booktabs}
\usepackage{float}
\usepackage{bm}
\newcommand{\tabincell}[2]{\begin{tabular}{@{}#1@{}}#2\end{tabular}}
\newcommand{\sysname}{NAR4TSP}

\usepackage{threeparttable}
\usepackage{tabularx}
\usepackage{diagbox}

\usepackage{subfigure}
\usepackage{array}
\usepackage{textcomp}
\usepackage{stfloats}
\usepackage{url}
\usepackage{verbatim}
\usepackage{graphicx}
\usepackage{cite}
\usepackage{soul}
\usepackage{soul, color, xcolor}
\soulregister{\cite}7 
\soulregister{\citep}7 
\soulregister{\citet}7 
\soulregister{\ref}7 
\soulregister{\pageref}7 

\begin{document}

\title{Reinforcement Learning-based Non-Autoregressive Solver for Traveling Salesman Problems}

\author{Yubin~Xiao, Di~Wang,~\IEEEmembership{Senior~Member,~IEEE}, Boyang~Li, Huanhuan~Chen,~\IEEEmembership{Senior~Member,~IEEE}, Wei~Pang, Xuan~Wu, Hao~Li, Dong~Xu, Yanchun~Liang, and You~Zhou$^\star$
\thanks{This work is supported by the Jilin Provincial Department of Science and Technology Project (20230201083GX, 20240101369JC, and 20240302086GX), the Nanyang Associate Professorship, and the National Research Foundation Fellowship (NRF-NRFF13-2021-0006), Singapore. Any opinions, findings and conclusions or recommendations expressed in this material are those of the authors and do not reflect the views of the funding agencies. (\emph{$^\star$Corresponding author: You Zhou})}
\thanks{Yubin Xiao, Xuan Wu, and You Zhou are with the Key Laboratory of Symbolic Computation and Knowledge Engineering of Ministry of Education, College of Computer Science and Technology, Jilin University, Changchun, 130012, China (e-mail: zyou@jlu.edu.cn).}
\thanks{Di Wang is with the Joint NTU-UBC Research Centre of Excellence in Active Living for the Elderly, Nanyang Technological University, 639798, Singapore.}
\thanks{Boyang Li is with the School of Computer Science and Engineering, Nanyang Technological University, 639798, Singapore.}
\thanks{Huanhuan Chen is with the School of Computer Science, University of Science \& Technology of China, Hefei, 230026, China.}
\thanks{Wei Pang is with the School of Mathematical and Computer Sciences, Heriot-Watt University, Edinburgh EH14 4AS, U.K.}
\thanks{Hao Li is with College of Computer, National University of Defense Technology, Changsha, Hunan, 410073, China.}
\thanks{Dong Xu is with the Department of Electrical Engineering and Computer Science, Bond Life Sciences Center, University of Missouri, Columbia, USA.}
\thanks{Yanchun Liang is with the School of Computer Science, Zhuhai College of Science and Technology, Zhuhai 519000, China.}
}



\maketitle

\begin{abstract}
The Traveling Salesman Problem~(TSP) is a well-known combinatorial optimization problem with broad real-world applications. Recently, neural networks have gained popularity in this research area because as shown in the literature, they provide strong heuristic solutions to TSPs. Compared to autoregressive neural approaches, non-autoregressive~(NAR) networks exploit the inference parallelism to elevate inference speed but suffer from comparatively low solution quality. In this paper, we propose a novel NAR model named \sysname, which incorporates a specially designed architecture and an enhanced reinforcement learning strategy. To the best of our knowledge, \sysname{} is the first TSP solver that successfully combines RL and NAR networks. The key lies in the incorporation of NAR network output decoding into the training process. \sysname{} efficiently represents TSP encoded information as rewards and seamlessly integrates it into reinforcement learning strategies, while maintaining consistent TSP sequence constraints during both training and testing phases. Experimental results on both synthetic and real-world TSPs demonstrate that \sysname{} outperforms five state-of-the-art models in terms of solution quality, inference speed, and generalization to unseen scenarios.
\end{abstract}



\section{Introduction}
\label{section:Introduction}
\IEEEPARstart{T}HE Traveling Salesman Problem~(TSP) is one of the most well-known Combinatorial Optimization Problems~(COPs) \cite{Garey1990} and can be defined on a connected graph with non-negative edge weights. A TSP solution is a Hamiltonian cycle that minimizes the sum of edge weights in the path. Here, the Hamiltonian cycle is a path that starts and ends at the same node and visits every other node exactly once. TSP has a wide range of real-world applications, such as maritime transportation \cite{Christiansen2004} and facility placement \cite{Charikar2001}. Due to the NP-hard nature of TSP, many approximate and heuristic algorithms  \cite{Onizawa2022} have been developed over the years. 

Recently, advances in deep learning have led researchers to develop neural networks~(NNs) as a viable solver for TSPs \cite{Zhang2022, Shao2023, Zhang2023, Gao2023, Wang2024, Prashant2023}. While theoretical guarantees for such networks remain elusive, they tend to produce near-optimal solutions in practice, with faster inference speed and better generalization compared to conventional TSP solvers \cite{Bengio2021, Wu2024}. However, most existing NN solutions adopt the autoregressive~(AR) approach, which produces nodes on the Hamiltonian cycle one at a time. The sequential nature of these networks poses a fundamental limit on the speed of the algorithm. On the other hand, non-autoregressive~(NAR) networks \cite{Nowak2018, Joshi2019} feature fast inference, but tend to produce comparatively low-quality solutions, especially when the test problems differ in size and distribution from those encountered during training\cite{Joshi2019}.

This paper focuses on NAR networks for solving TSPs. We attribute the limited popularity of NAR models to the following three key limitations. Firstly, conventional NAR models adopt supervised learning~(SL) for training, requiring predetermined optimal solutions for numerous TSP instances as training data through exact TSP solvers (e.g., Concorde \cite{Applegate2007}), which is computationally expensive. Secondly, the SL-trained models are vulnerable to overfitting, leading to poor generalization performance on other TSPs \cite{Joshi2019}. Thirdly, their training process is not completely consistent with the inference process, resulting in less competitive performance. Specifically, they relax the sequential/tour constraint of TSP during training, estimating the importance of each edge in parallel. However, they impose such constraints during inference, forming feasible TSP solutions by searching for essential edges.

To address the aforementioned limitations of conventional NAR models, we propose a novel NAR model called NAR4TSP with several improvements. First, we propose a modified architecture for the graph neural network (GNN), aiming to facilitate the discovery and optimization of Hamiltonian cycles by extracting information from graphs represented by TSPs. Specifically, we introduce a learnable pointer that indicates the starting node, whereas existing techniques \cite{Nowak2018, Joshi2019} always start at the first node. Our proposed GNN takes the node coordinates and distances between nodes as inputs, and outputs scores for all edges and the starting-node pointer. Secondly, to ensure the consistency of the sequential constraints of TSP being applied in both the training and inference phases, which is crucial for producing high-quality solutions in AR models \cite{Kool2019, Bresson2021, Jung2023, Yang2023}, we impose the sequential constraints of TSP during training~(i.e., not only during inference). Specifically, we decode the output of the GNN with such constraints to produce a TSP solution, which is then assessed and used as a reward for the self-critical reinforcement learning (RL) strategy \cite{Rennie2017}. Thirdly, we enhance the vanilla RL strategy for training our NAR models to perform better with fewer computing resources~(see Section~\ref{training} for more details). Specifically, we leverage the one-shot nature of the NAR networks, replacing the two modules used in the vanilla RL strategy with a single module. Through RL training, \sysname{} addresses the limitations of using SL in conventional NAR models, i.e., it does not require the expensively produced ground-truth labels. 

It is worth noting that combining NAR network and RL technology appears challenging due to their contrasting nature. The NAR network produces parallel outputs, whereas the RL process demands a step-by-step implementation. However, we introduce a decoding process that progressively transforms the NAR network's outputs into the TSP solutions, thus, seamlessly integrating RL into our approach. To the best of our knowledge, \sysname{} is the first NAR model trained using RL for solving TSPs.


We evaluate \sysname{} in three aspects, namely solution quality, inference latency, and generalization to unseen scenarios. We adopt thirteen neural network baselines \cite{Kool2019,Bresson2021,Jung2023,Vinyals2015,Bello2017,Khalil2017,Deudon2018,Joshi2019,Wu2022, Costa2021, Yang2023, Fellek2024, liu2024}, and compare them with \sysname{} on TSP50~(TSP with 50 nodes) and TSP100~(TSP with 100 nodes). Despite the one-shot nature of NAR decoding \cite{Joshi2022}, \sysname{} produces competitive solutions while achieving significantly shorter inference time. To evaluate the inference speed of \sysname, we measure its inference time and compare it to that of five state-of-the-art~(SOTA) NN-based models \cite{Joshi2019, Kool2019, Bresson2021, Jung2023, Yang2023} using TSPs of different sizes and beam search with different widths. The experimental results show that \sysname{} outperforms the other models in terms of inference speed, especially when the number of nodes and the width of beam increase. Finally, we evaluate the out-of-domain generalization of \sysname{} to TSP instances of different sizes from that of training data and $35$ real-world instances. The experimental results indicate that \sysname{} has a superior generalization ability comparing against the other SOTA models \cite{Joshi2019, Kool2019, Bresson2021, Jung2023, Yang2023}. In addition to result comparisons, we demonstrate the feasibility of implementing \sysname{} in an end-to-end manner and the effectiveness of \sysname{} for solving TSPs through the visualization of the decoding process and the overall path planning of \sysname, respectively. Furthermore, we demonstrate the effectiveness of extending NAR4TSP to solve other COPs, such as the Manhattan TSP and Capacitated Vehicle Routing Problem~(CVRP).

The key contributions of this research work are as follows:
\begin{itemize}
\item[$\bullet$]To the best of our knowledge, we propose the first NAR model trained using RL for solving TSPs, eliminating the need for expensive ground-truth labels used to train the conventional SL-based NAR models.

\item[$\bullet$]We construct a novel GNN as the backbone architecture of \sysname{} and incorporate sequential constraints of TSP during training to elevate the model performance.

\item[$\bullet$]We show the excellent performance of \sysname, especially in inference speed and generalization ability, by conducting extensive experiments.

\item[$\bullet$]We demonstrate \sysname's ease in implementation and effectiveness by visualizing its decoding process and overall path planning, respectively.

\end{itemize}
%
\begin{table}[!t]
  \centering
  \caption{End-to-end NN-based models proposed to solve TSPs}
  \resizebox{1\columnwidth}{!}{
    \begin{tabular}{c|p{0.19\textwidth}p{0.22\textwidth}}
    \toprule
    \diagbox[width=0.12\textwidth,trim=l]{\textbf{Training:}}{\textbf{Decoding:}} & \multicolumn{1}{c}{AR} & \multicolumn{1}{c}{NAR} \\
    \midrule
    \multicolumn{1}{c|}{\multirow{2}{*}{SL}} & \multicolumn{1}{c|}{\multirow{2}{*}{Vinyals \emph{et al.} \cite{Vinyals2015}}} & Nowak \emph{et al.} \cite{Nowak2018}; Joshi \emph{et al.} \cite{Joshi2019}; Xiao \emph{et al.} \cite{Xiao2024};  Liu \emph{et al.} \cite{liu2024}\\
    \cmidrule{2-3}
    \multicolumn{1}{c|}{\multirow{6}{*}{RL}} & Kool \emph{et al.} \cite{Kool2019}; Bresson and Laurent \cite{Bresson2021}; Jung \emph{et al.} \cite{Jung2023}; Bello \emph{et al.} \cite{Bello2017};  Khalil \emph{et al.} \cite{Khalil2017}; Deudon \emph{et al.} \cite{Deudon2018}; Yang \emph{et al.} \cite{Yang2023}; Fellek \emph{et al.} \cite{Fellek2024}
    & \multicolumn{1}{|c}{\multirow{6}{*}{\sysname~(our work)}} \\
    \bottomrule
    \end{tabular}}%
  \label{outline}%
\end{table}%

\section{Related Work}
\label{section:Relate work}
In this section, we review the conventional algorithms for solving TSPs and the relevant NN-based TSP solvers.

\subsection{Conventional TSP Solvers}
Conventional algorithms for solving TSPs can be generally classified into exact and approximate methods, as well as heuristics. The first category, exemplified by Concorde \cite{Applegate2007}, is known to produce optimal TSP solutions through integer programming and optimization algorithms, including Cutting Planes and Branch-and-Bound \cite{Padberg1991, Applegate2003, Bellman1962}. The second category typically employs linear programming and relaxation techniques to produce TSP solutions with guaranteed quality \cite{Hochba1997}. The third category, represented by LKH3 \cite{Helsgaun2017} and its predecessor LKH2 \cite{Helsgaun2009}, is based on the $k$-opt heuristics and has been shown to produce solutions almost as good as those generated by Concorde \cite{Kool2022}. Within this category, evolutionary computation algorithms (e.g., Ant Colony Optimization (ACO)) \cite{Deng2022, Yang2020} and local search-based algorithms (e.g., Variable Neighborhood Search (VNS)) have been widely applied to solve TSPs. Additionally, several (supervised and unsupervised) learning-based methods \cite{Faigl2018, Faigl2018a, Wang2017, Decker2022, Franti2021} have been proposed to improve the computational efficiency and solution quality of existing heuristics, demonstrating excellent performance on certain benchmarking datasets such as TSBLIB.

However, conventional TSP solvers may not be suitable for real-time TSP tasks due to their inability to produce high-quality solutions within a short time frame and to solve multiple TSP instances simultaneously \cite{Li2021}.

\subsection{Neural Network-based TSP Solvers}\label{NNsolver}
With the recent advancements in deep learning, numerous NN-based models have been proposed to solve TSPs \cite{Zhou2023, Li2023, Wang2024, Wang2024a}, which can be broadly categorized into neural improvement type and end-to-end learning type. Neural improvement methods \cite{Andoni2023} commence with initial solutions and employ specific deep learning techniques (e.g., a pretrained NN) to guide or assist heuristics to iteratively improve the solution quality \cite{Li2023}. In line with this research, local search (e.g., NeuroLKH \cite{Xin2021a}) and evolutionary computation (e.g., DeepACO \cite{ye2023}) algorithms are often utilized. It is worth noting that we focus on end-to-end learning methods for solving TSPs in this study, and exclude neural improvement models because they typically have longer inference latency \cite{Kool2022}. End-to-end learning model can be generally categorized into AR or NAR according to their model decoding methods, and SL or RL according to their model training methods. We provide an overview of the difference between the existing end-to-end NN-based models and our work in Table~\ref{outline}. 


We first review the pioneering NN-based approaches used to solve TSPs. Vinyals \emph{et al.} \cite{Vinyals2015} proposed a sequence-to-sequence SL model named Pointer Network~(PtrNet) to solve TSPs. PtrNet takes node coordinates as its input, employs the attention mechanism and generates TSP solutions step-by-step, using the optimal solutions produced by Concorde as the ground-truth labels. While PtrNet provides a novel perspective for solving combinatorial optimization~(CO) problems, its unsatisfactory performance and high cost of label production make it challenging to be applied in practical settings \cite{Li2021}. Compared to SL models, which require optimal solutions, RL is a more elegant alternative in the absence of ground-truth labels. Moreover, RL has demonstrated robust learning and optimized decision-making ability in various domains such as AlphaGo Zero \cite{Silver2017}. Thus, majority existing AR approaches use RL for training, lifting the prerequisite of optimal solutions \cite{Bello2017, Khalil2017, Deudon2018}. With Transformer setting new performance records in various applications \cite{Vaswani2017}, numerous Transformer-based TSP models have been proposed, and demonstrate excellent performance \cite{Kool2019, Bresson2021, Yang2023, Jung2023, liu2024, Fellek2024, Xiao2024A}.



However, the type of step-by-step solution generation method used by AR models has a natural disadvantage in terms of inference speed when compared to the NAR approaches \cite{Ran2021}. Additionally, due to the sequential nature of AR models, beam search suffers from diminishing returns with respect to the beam size and shows limited search parallelism \cite{Koehn2017}. To address these issues, several NAR models have been developed to improve the inference speed when solving TSPs \cite{Joshi2019, Nowak2018, Xiao2024}. These models regard TSP as a link prediction problem that has been studied in various relevant research fields \cite{Xiao2020}. They use SL to train their models to estimate the importance of each edge belonging to the optimal solution. For instance, Joshi \emph{et al.} \cite{Joshi2019} utilized a Graph Convolutional Network~(GCN)-based TSP solver trained with SL. The model takes a TSP instance as a graph for model input and directly outputs a heat map that represents the importance of edges. It is worth noting that during training, the model relaxes the sequential constraint of TSP to obtain the edge's importance by minimizing the binary cross-entropy loss between the adjacency matrix corresponding to the optimal solution and the heat map. While during inference, the model imposes such constraints by using greedy search or beam search on the heat map to obtain a feasible TSP solution.


Despite the significant improvement in inference speed offered by NAR approaches, the solution quality of these models is often suboptimal \cite{Joshi2019}. In fact, existing NAR models generally fail to outperform most of the afore-reviewed AR models \cite{Ma2023}. As such, we deem there is a pressing need to improve the existing models further to achieve high-quality solutions with low inference latency.
\begin{figure*}[!t]
	\centering
	\includegraphics[width=2\columnwidth]{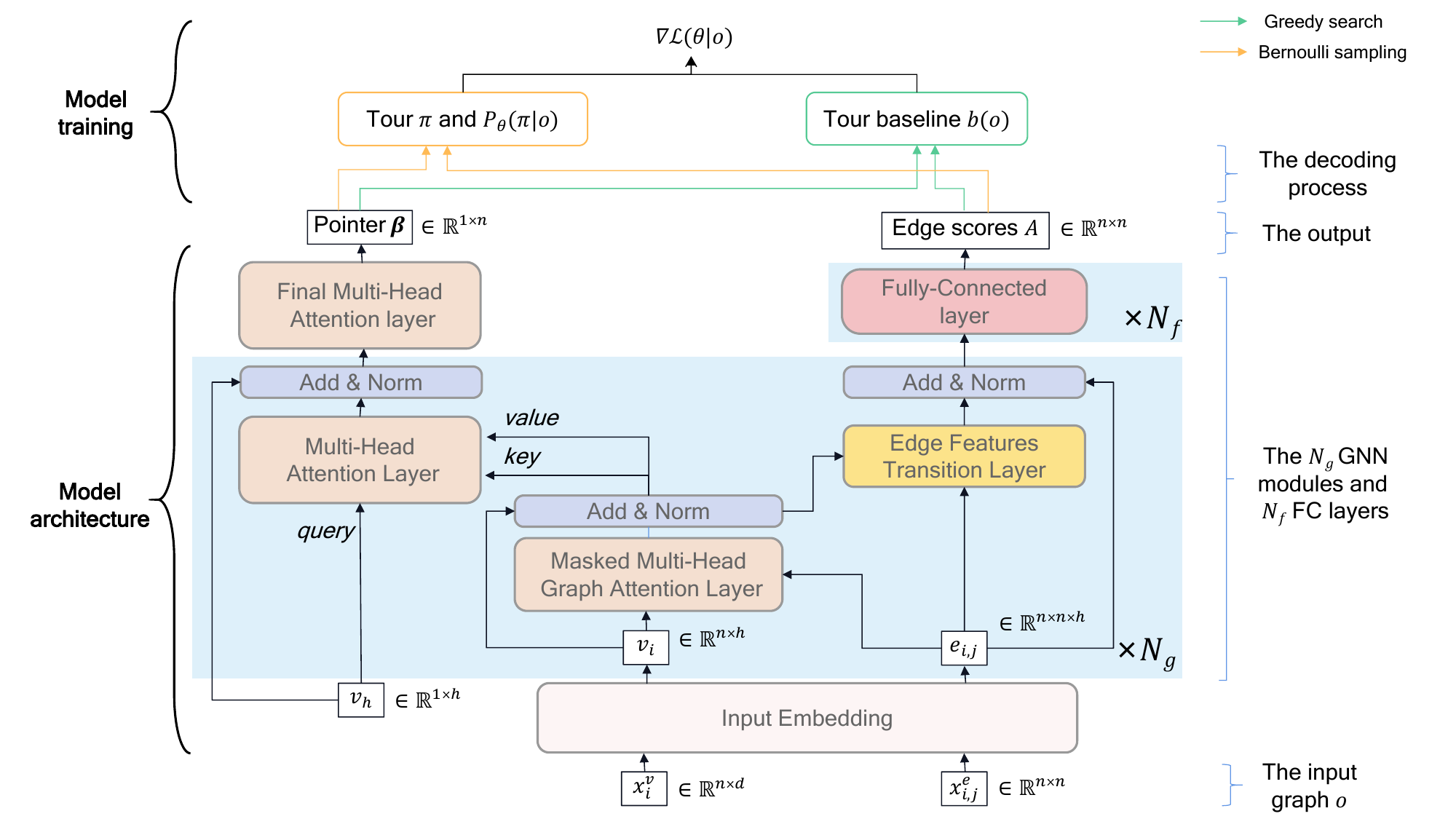}
	
	\caption{The pipeline of \sysname. Taking a TSP instance $o$ with the node coordinate $x^v_i$ and the Euclidean distance between nodes $x^{e}_{i,j}$ as inputs, the model starts by processing the information through a linear layer, transforming it into node features $v_i$ and edge features $e_{i,j}$. The model then interacts with a randomly initialized learnable starting symbol $v_h$ via $N_g$ GNN modules, and finally outputs a starting-node pointer $\bm{\beta}$ and a matrix~$A$ of edge scores. The output $\bm{\beta}$ and $A$ are subsequently decoded into a feasible TSP tour through sampling or greedy search. The solution is obtained in a one-shot, NAR manner, and its quality is treated as a reward and optimized by an enhanced RL strategy.}
	
	\label{fig1}
\end{figure*}

\section{\sysname}
\label{section:\sysname}
In this section, we present a novel NAR model named \sysname, which is designed to better solve TSPs. As shown in Fig.~\ref{fig1}, \sysname{} adopts a unique GNN architecture to output a starting-node pointer and a matrix of edge scores. It then decodes the GNN's output to produce feasible TSP solutions for model training. We detail the TSP setting, the architecture, the decoding process, the enhanced RL strategy, and the complexity analysis of \sysname, and potential adoption of \sysname~in solving other COPs in the following subsections, respectively.

\subsection{TSP Setting}
Our research focuses on the most fundamental Euclidean TSP due to its importance and prevalence in various domains \cite{Nowak2018, Joshi2019, Deudon2018, Kool2019, Bresson2021}. We present a TSP instance as a graph $G=(V,E)$ with $n$ nodes in the node set $V$, and $n\times n$ edges in the edge set $V$, where node $x^{v}_i \in \mathbb{R}^{n\times d}$ denotes the $d$-dimensional node coordinates and edge $x^{e}_{i,j}\in \mathbb{R}^{n\times n}$ denotes the Euclidean distance between nodes $x^{v}_i$ and $x^{v}_j$.

We define a TSP tour as a permutation of $n$ nodes denoted by $\pi = \{\pi_1,\pi_2,...,\pi_n\}$, where $\pi_i\neq \pi_j, \forall i\neq j$. The length of a TSP tour $\pi$ is defined as follows:
\begin{equation}\label{eq1}
	\operatorname{L(\pi)}=x^{e}_{\pi_n, \pi_1} + \sum_{i=1}^{n-1}x^{e}_{\pi_i, \pi_{i+1}},
\end{equation}
where $x^{e}_{\pi_i, \pi_j}$ denotes the Euclidean distance between nodes $\pi_i$ and $\pi_j$. In this study, we use the term TSP to specifically refer to the Euclidean TSP for clarity and convenience.

\subsection{Architecture of \sysname}
\label{section:architecture}
As shown in Fig.~\ref{fig1}, the architecture of \sysname{} consists of three key components, namely 1)~a linear layer for input embedding; 2)~multiple GNN modules for facilitating information transfer among nodes, edges and the starting symbol, and ultimately for outputting the starting-node pointer; and 3)~several fully-connected~(FC) layers for outputting edge scores. For the ease of subsequent descriptions, we denote the number of GNN modules and FC layers as $N_g$ and $N_f$, respectively. 

It is worth noting that the TSP solution forms a Hamiltonian cycle, allowing any node to be selected as the starting node. Nevertheless, for guaranteeing the production of a valid solution, the implementation of a mask mechanism becomes necessary. This mechanism masks the probability of pointing to previously selected nodes, thereby preventing nodes from being revisited. Consequently, the resulting probability distribution of earlier node selections shows a more significant dispersion compared to the later selections, because the probability of selecting previously visited nodes at each selection is masked to zero. Therefore, the selection of the first node presents the most dispersed probability distribution, underscoring the challenge and significance of the selection of the starting node.

Drawing inspiration from natural language processing research, where special symbols~(e.g., ``$\langle bos\rangle$'') serve various purposes, such as marking the beginning of a sentence. We design the pointer in the GNN modules for the purpose of initiating the decoding at the most promising starting position to produce TSP solutions.  This approach differs from conventional NAR models, which typically take the node indexed at 1 as the starting node \cite{Nowak2018, Joshi2019}. By utilizing the pointer mechanism, we establish a framework to incorporate the sequential constraint of TSP into the decoding process, thus applying the consistent sequential constraints in both training and inference phases, unlike the afore-reviewed NAR models \cite{Nowak2018, Joshi2019}~(see Section~\ref{NNsolver}). To the best of our knowledge, this novel GNN architecture, featuring a starting-node pointer output, represents a pioneering architecture that leverages TSP representations to guide the decoding process in NAR models. The effectiveness of this novel architecture is assessed later through three ablation studies~(see Section~\ref{section:effectiveness}).

For a given TSP graph, \sysname{} first linearly projects the input nodes $x^{v}_i \in \mathbb{R}^{n\times d}$ and edges $x^{e}_{i,j}\in \mathbb{R}^{n\times n}$ into $h$-dimensional features. Their initial embedding is as follows:
\begin{equation}\label{eq2}
  v_i^0=\bm{W_{v}}x_i^{v}+\bm{b_v},v_i^0\in\mathbb{R}^{n\times h},
\end{equation}
\begin{equation}\label{eq3}
e_{i,j}^0=\bm{W_{e}}x_{i,j}^e+\bm{b_{e}},e_{i,j}^0\in\mathbb{R}^{n\times n\times h},
\end{equation}
where $\bm{W}$ and $\bm{b}$ denote the learnable parameters of weights and biases of the GNN, respectively.

\sysname{} utilizes $N_g$ GNN modules to extract feature information from the graph. GNN is a specific type of NN that is well-suited for information interaction between nodes and edges in graphs, and its effectiveness has been demonstrated in prior studies \cite{Joshi2019}. In our work, each GNN module is used for three types of feature information transfer, namely nodes, edges, and the starting symbol. We detail these feature information transfer mechanisms in the subsequent paragraphs.
\subsubsection{\textbf{Node}} For the transfer of node feature information, we integrate the edge features with the node features using the vanilla multi-head graph attention module \cite{Velickovic2018}. Because it is trivial to extend single head to multi-head, we use a single head in all formulas for the convenience of subsequent descriptions, as done in \cite{Bresson2021}. As an initial step, we apply a linear transformation parameterized by a weight matrix to all nodes and edges, respectively. We then apply the LeakyReLU nonlinear activation function $\sigma_1$ and compute the importance of node $j$'s features to node $i$ as follows:
\begin{equation}\label{eq4}
\small
\lambda_{i,j}^{l}=\sigma_1(\bm{W_e^{l}}[\bm{W_{vv}^{l}}[v_i^{l-1}||v_j^{l-1}]||\bm{W_{ve}^{l}}e_{i,j}^{l-1}]),\lambda_{i,j}^{l}\in\mathbb{R}^{n\times n},
\end{equation}
\normalsize
\begin{equation}\label{eq5}
	\sigma_1(z)=\begin{cases}
		z, & \mbox{if } z>0, \\
		\delta z, & \mbox{otherwise},
	\end{cases}
\end{equation}
where $l\in\{1,2,3,\dots,N_g\}$ denotes the GNN module index, ${{||}}$ denotes the concatenation operator, and $\delta$ denotes the trade-off parameter. To facilitate the comparison of importance scores across different nodes, we apply the Softmax activation function $\sigma_2$ to normalize them, which is defined as follows:
\begin{equation}\label{eq6}
	\alpha_{i,j}^{l}=\sigma_2(\lambda_{i,j}^{l}\odot N_{i,j}),\alpha_{i,j}^{l}\in\mathbb{R}^{n\times n},
\end{equation}
\begin{equation}\label{eq7}
	\sigma_2(z_i) = \frac{e^{z_{i}}}{\sum_{j=1}^n e^{z_{j}}},\forall i\in\{1,2,\dots,n\},
\end{equation}
\begin{equation}\label{eq8}
	N_{i,j}=\begin{cases}
		1,  & \text{ if node}~j\in N_{set(i)}, \\
		-\infty,  & \text{ otherwise},
	\end{cases}
\end{equation}
where $\odot$ denotes the element-wise multiplication and $N_{i,j}$ denotes the mask on whether node $j$ belongs to the neighboring set~$ N_{set(i)}$ of node $i$. This approach allows GNN to selectively filter out non-neighboring node information and focus solely on computing the importance of nodes $j\in  N_{set(i)}$. It is worth noting that in the TSP graph, all nodes are connected, implying that the neighboring set $N_{set(i)}$ of each node encompasses all the other nodes except itself. To expedite the model's training process, we select the top $\frac{n}{k}$-nearest nodes as the neighbors of node $i$ if $n$ exceeds the threshold $k$. Finally, we compute the node features at the subsequent layer as follows:
\begin{equation}\label{eq9}
	v_i^{l}=\operatorname{BN}((\sum _{j}^{n} \alpha_{i,j}^{l}\odot v_j^{l-1})+v_i^{l-1}),v_i^{l}\in\mathbb{R}^{n\times h},
\end{equation}
where $\operatorname{BN}$ denotes Batch Normalization \cite{Ioffe2015}.

\subsubsection{\textbf{Edge}} The conventional approach for transferring edge feature information aggregates adjacent edges' features to update each edge's attributes \cite{Wang2021}. However, in the TSP graph, all nodes are connected to each other. Hence, it is challenging to efficiently identify the adjacent edges. One of the promising strategies to address this issue is to utilize the information from the nodes at both ends of each edge to update its features. To accomplish this, we employ a single-layer feedforward NN with a trainable weight matrix and a bias term, which operates on all nodes, and compute their contribution $C$ to the edge's features as follows:
\begin{equation}\label{eq10}
	C_{i,j}^{l}=v1_i^{l}\oplus v2_j^{l}, C_{i,j}^{l}\in\mathbb{R}^{n\times h\times h},
\end{equation}
\begin{equation}\label{eq11}
	v1_i^{l}=\bm{W_{e1}^{l}}v_i^l + \bm{b1_{i}^l}, v1_i^{l}\in \mathbb{R}^{n\times h\times 1},
\end{equation}
\begin{equation}\label{eq12}
	v2_j^{l}=\bm{W_{e2}^{l}}v_j^l + \bm{b2_{j}^l}, v2_j^{l}\in \mathbb{R}^{n\times 1\times h},
\end{equation}
where $\oplus$ denotes the addition operation with broadcasting. After computing each node's contribution to the edge features, we aggregate these contributions and apply the Sigmoid activation function $\sigma_3$ to compute the edge features at the subsequent layer as follows:
\begin{equation}\label{eq13}
\footnotesize
	e_{i,j}^{l}=\operatorname{BN}(\sigma_3(C_{i,j}^{l}+\bm{W_{ee}^l}e_{i,j}^{l-1} + \bm{b_{i,j}^l})+e_{i,j}^{l-1}),e_{i,j}^l\in\mathbb{R}^{n\times n\times h},
\end{equation}
\normalsize
\begin{equation}\label{eq14}
	\sigma_3(z) = \frac{1} {1 + e^{-z}}.
\end{equation}

\begin{figure*}[!t]
\centering
\includegraphics[width=2\columnwidth]{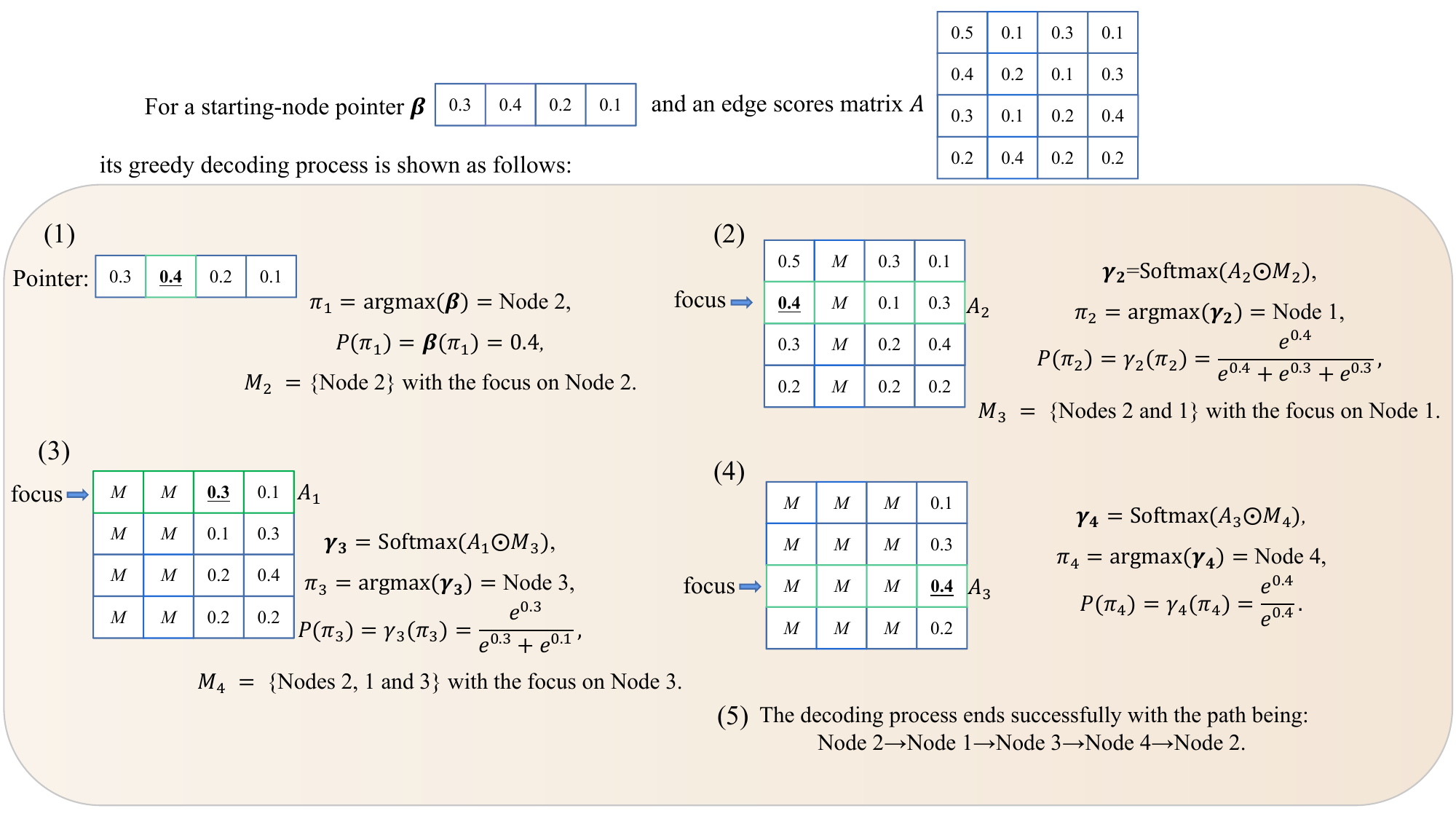}
\caption{An illustration of the decoding process of \sysname{} using greedy search, assuming there are four nodes in the graph.}

\label{fig2}
\end{figure*}
\subsubsection{\textbf{Starting symbol}} For the transfer of the starting symbol feature information, we introduce a virtual node $v_h\in \mathbb{R}^{1\times h}$, which is randomly initialized and does not belong to the node set~$V$ (but virtually match to one node in $V$). This virtual node employs the self-attention mechanism \cite{Vaswani2017} to learn and ultimately identify the most promising starting node in TSP solutions. Self-attention is a powerful technique used in conventional models \cite{Bresson2021} to map a query and a set of key-value pairs to an output. In this study, we leverage self-attention to interactively update the starting symbol's features by aggregating all features of each node. Specifically, we define the query as the features of the starting symbol $v_h$ and the key and value as the features of node $v$, as follows:
\begin{equation}\label{eq15}
	q^{l}=\bm{W_q^{l}}v_h^{l-1}, q^{l}\in\mathbb{R}^{1\times h},
\end{equation}
\begin{equation}\label{eq16}
	K_i^{l}=\bm{W_k^{l}}v_i^{l}, K^{l}\in\mathbb{R}^{n\times h},
\end{equation}
\begin{equation}\label{eq17}
	V_i^{l}=\bm{W_v^{l}}v_i^{l}, V^{l}\in\mathbb{R}^{n\times h},
\end{equation}
where $v_h^0$ denotes the initial features of the starting symbol, which is equal to $v_h$. The starting symbol's features at the subsequent layer are given as:
\begin{equation}\label{eq18}
	v_h^{l}=\operatorname{BN}(\sum_{i}^{n}\sigma_1(q^{l}{K_i^l}^T)\odot V_i^{l}+v_h^{l-1}),v_h^{l}\in\mathbb{R}^{1\times h}.
\end{equation}
In the last layer of the GNN, the starting symbol acquires the probability of each node in $V$ being the starting node by querying all nodes as follows:
\begin{equation}\label{eq19}
\bm{\beta}=\sigma_1(qK^T), \bm{\beta}\in\mathbb{R}^{1\times n},
\end{equation}
\begin{equation}\label{eq20}
q=\bm{W_q^{N_g}}v_h^{N_g-1}, q^{N_g}\in\mathbb{R}^{1\times h},
\end{equation}
\begin{equation}\label{eq21}
K=\bm{W_k^{N_g}}v^{N_g}, K\in\mathbb{R}^{n\times h},
\end{equation}
where $\bm{\beta}$ denotes the probability distribution of $n$ nodes as the starting node in TSP solutions. We use $\bm{\beta}$ to determine the pointer indicating where to initiate the decoding (see (\ref{eq26})) and let it be one model output.

The final component of \sysname{} comprises $N_f$ FC layers, which transform the edge features to prediction scores for the edges' likelihood of being selected in the TSP tour. Specifically, we use the edge features $e_{i,j}^{N_g}$ as the embedding of the edge at the first FC layer as follows:
\begin{equation}\label{eq22}
	A^0=e_{i,j}^{N_g}, A^0\in\mathbb{R}^{n\times n\times h}.
\end{equation}
We then use the ReLU activation function $\sigma_4$ to compute the edge embedding at the subsequent layer as follows:
\begin{equation}\label{eq23}
	A^{l}=\sigma_4(\bm{W_f^l}A^{l-1}+\bm{b_f^l}), A^{l}\in\mathbb{R}^{n\times n\times h},
\end{equation}
\begin{equation}\label{eq24}
	\sigma_4(z) = \operatorname{max}(0, z),
\end{equation}
where $l\in\{1,2,3,\dots,N_f-1\}$ denotes the FC layer's index. The forward pass of the last layer is defined as follows:
\begin{equation}\label{eq25}
A=\bm{W_f^{N_f}}A^{N_f-1}+\bm{b_f^{N_f}},A\in\mathbb{R}^{n\times n\times 1}.
\end{equation}
Finally, the model compresses the third axis to obtain the two-dimensional matrix of edge scores $A\in\mathbb{R}^{n\times n}$. This matrix $A$ denotes the tightness of links between $n$ nodes and is the other model output~(other than the starting-node pointer).

\subsection{Decoding Process of \sysname}
In the decoding process of the conventional NAR models \cite{Nowak2018, Joshi2019}, the starting position for inference is typically set as the node indexed at $1$. In contrast, we deem that selecting the starting position in a data-driven manner is more practical than always using a predetermined index as the starting position. This is the motivation of our design of the starting symbol in the GNN. Note that the starting position in this context refers to the index of a node, from which the model should start the decoding process. It does not mean that the tour has to start from this node. After the decoding process completes, one may always start the TSP tour from the node indexed at $1$ or from any other predetermined node.

Fig.~\ref{fig2} shows an example of the decoding process of \sysname{} using greedy search. To obtain a feasible TSP solution $\pi$ from the output of \sysname, i.e., a matrix of edge scores $A$ and a starting-node pointer $\bm{\beta}$, we incorporate the sequential constraints of TSP to decode them using the greedy search algorithm. Specifically, we start by generating the starting node as follows:
\begin{equation}\label{eq26}
	\pi_1=\operatorname{argmax}(\bm{\beta}).
\end{equation}
Then, we apply the Softmax activation function $\sigma_2$ to normalize the edge scores, representing the tightness of the node links. We compute the probability distribution of the subsequent node by masking the probability of already visited nodes as follows:
\begin{equation}\label{eq27}
\small
\bm{\gamma_i}=\sigma_2 \left( { A_{i,j}}\odot\begin{cases}
	-\infty, & \text{ if node}~j \in  M_i, \\
	1, & \text{ otherwise},
\end{cases}\right ), \bm{\gamma_i}\in\mathbb{R}^{1\times n},
\end{equation}
\normalsize
where $i \in \{2, 3,\dots,n\}$, $ A_{i,j}$ denotes the element value in the $i$th row and $j$th column of $A$, representing the tightness of the link between nodes $i$ and $j$, and set $ M_i$ consists of all visited nodes, i.e., $\pi_1,\pi_2,\dots,\pi_{i-1}$. Finally, we obtain the sequence of the remaining nodes by adopting the greedy search algorithm as follows:
\begin{equation}\label{eq28}
\pi_i=\operatorname{argmax}(\bm{\gamma_i}).
\end{equation}

More advanced sampling methods, such as beam search, are commonly used in solving TSPs \cite{Nowak2018, Bresson2021}. Beam search is a breadth-first search method that operates within a limited width, denoted as $B$. This method commences its exploration from initial candidates~(e.g., the starting node) and incrementally expands the tours by evaluating $B$ potential successors. It selectively retains the top-$B$ tours with the highest scores at each step, determined by their cumulative logarithmic probabilities. Greedy search is a special case of beam search with the width of $1$. From the perspective of dynamic programming, setting $B$ to $n\cdot 2^n$ guarantees optimal results \cite{Kool2022}, but selecting a small value of $B$ greatly alleviates computational cost, albeit with possible degradation on performance. When using beam search, AR models exhibit limited search parallelism, because they require multiple probability distribution computations in each decoding stage \cite{Kool2019, Bresson2021}. In contrast, NAR models such as \sysname{} only perform search in the model output without multiple computations, significantly reducing the inference time \cite{Nowak2018, Joshi2019}. The time efficiency of \sysname{} is further demonstrated with experimental results in Section~\ref{section:evaluation}.

\subsection{Enhanced Reinforcement Learning Strategy of \sysname}
\label{training}
RL-based TSP solvers aim to minimize the overall length of all edges in a TSP tour. For a given TSP instance $o$, the loss function used for training is the overall tour length $\operatorname{L(\pi)}$ of the current TSP solution:
\begin{equation}\label{eq29}
\mathcal{L}(\theta|o)=\mathbb{E}_{P_{\theta}}[\operatorname{L}(\pi)],
\end{equation}
where $P_{\theta}$ denotes the policy parameterized by $\theta$ and the gradient of~(\ref{eq29}) can be computed as follows based on the well-known REINFORCE algorithm \cite{Williams1992}:
\begin{equation}\label{eq30}
\nabla\mathcal{L}(\theta|o)=\mathbb{E}_{P_{\theta}}[\operatorname{L}(\pi)-\operatorname{L}(b(o))]\nabla \operatorname{log}\, P_{\theta}(\pi|o),
\end{equation}
where $P_{\theta}(\pi|o)$ denotes the probability of the TSP tour $\pi$ based on the policy $P_\theta$ and $b(o)$ denotes the tour baseline for the TSP instance $o$.  

A good baseline can reduce the gradients' variance, thereby accelerate the model's learning process \cite{Williams1992}. Conventional models \cite{Kool2019, Bresson2021} employ the self-critical RL strategy \cite{Rennie2017} consisting of two sub-modules with an identical architecture and initial parameters. Specifically, one sub-module predicts the tour $\pi$ for training, while the other estimates the baseline~$b(o)$. During training, these two sub-modules use different sampling methods: Bernoulli sampling for the prediction sub-module and greedy search for the baseline sub-module. After multiple training iterations, if the prediction sub-module outperforms the baseline sub-module, the latter adopts the parameters of the former to obtain a better baseline~$b(o)$.

Although we employ RL to train our NAR model as well, different from the commonly adopted self-critical RL strategy, our proposed RL strategy in \sysname{} takes advantage of the one-shot nature of NAR models. Specifically, our enhanced RL strategy only uses one training sub-module that directly utilizes both Bernoulli sampling and greedy search on the model's output to obtain the prediction $\pi$ and the baseline $b(o)$, respectively. This approach eliminates the need for an additional sub-module, reducing required computational resources~(see Section~\ref{section:effectiveness}). In the following paragraphs, we present a comprehensive and detailed introduction to our enhanced RL strategy.

\subsubsection{\textbf{State}} The state $s_t$, where $t\in\{0,1,\dots,n \}$, is defined as the set of all previously visited nodes:
\begin{equation}\label{eq31}
s_t=
\begin{cases}
\{\varnothing\}, & \text{if } t=0, \\
\{\pi_1, \pi_2, \dots, \pi_{t}\}, & \text{otherwise}.
\end{cases}
\end{equation}

\subsubsection{\textbf{Action}} The action $a_t$ is defined as the node to be selected at time $t$:
\begin{equation}\label{eq32}
a_t=
\begin{cases}
\pi_{t+1}, & \text{if } t<n, \\
\pi_{1}, & \text{otherwise}.
\end{cases}
\end{equation}

\subsubsection{\textbf{Reward}} The reward $r(s_t,a_t)$ is defined as the negative cost (i.e., length in this paper, see (\ref{eq1})) incurred from taking action $a_t$ from state $s_t$:
\begin{equation}\label{eq33}
r(s_t,a_t)=
\begin{cases}
0, & \text{if } t=0, \\
-x^{e}_{a_{t-1}, a_{t}}, & \text{otherwise}.
\end{cases}
\end{equation}

\subsubsection{\textbf{Policy}} The policy $P_{\theta}(s_t)$ generates a probability distribution for the unvisited nodes based on the state $s_t$. It is worth noting that our policy differs from that of the conventional RL-based AR models \cite{Kool2019, Bresson2021} which requires interactions with the NN's weight each time when the probability distribution for the next node is obtained, i.e., each node is generated using a computationally intensive NN. Our policy only needs to access the GNN's weight once, retrieve the output pointer and the edge scores matrix at each stage~$t$, and subsequently determine which node to visit next until a complete TSP solution is produced. We deem both advantageous properties of \sysname, namely lesser number of RL sub-modules in use and much lesser number of required interactions between the policy and the external NNs, the key enhancement of our RL strategy used in \sysname. Upon visiting all nodes, the policy is then updated by maximizing the cumulative rewards $\sum_{t=0}^{n}r(s_t, a_t)$, i.e., $-\operatorname{L}(\pi)$~(see~(\ref{eq29})).

\begin{algorithm}[!t]
	\caption{The training process of \sysname.}
	\begin{algorithmic}[1]
		\REQUIRE the number of epochs $n_e$, steps per epoch $n_s$,  validation set size $n_v$, and batch size $n_{bs}$
		\STATE Initialize $\theta$, set $L_{tmp}\gets\infty$, and obtain validation set $o_{\textit{va}}$ by randomly generating $n_v$ TSP instances
		\FOR {$epoch$ in $1,...,n_e$}
		\FOR {$step$ in $1,...,n_s$}
		\STATE $o_{\textit{tr}}\gets$ Randomly generate $n_{bs}$ TSP instances
		\STATE $\bm{\beta}$, $A \gets$ \textbf{model} ($o_{\textit{tr}}, \theta$)
		\STATE $\pi(o_{\textit{tr}}), P_{\theta}(\pi|o_{\textit{tr}})\gets$ Bernoulli sample ($\bm{\beta}$, $A$)
		\STATE $b(o_{\textit{tr}})\gets$ Greedy search ($\bm{\beta}$, $A$)
		\STATE $\nabla\mathcal{L}\gets$ ($\operatorname{L}(\pi(o_{\textit{tr}})), \operatorname{L}(b(o_{\textit{tr}})), P_{\theta}(\pi|o_{\textit{tr}})$) (see~(\ref{eq36}))
		\STATE $\theta\gets$ Adam($\theta, \nabla\mathcal{L}$)
		\ENDFOR
		\STATE $\bm{\beta}$, $A \gets$ \textbf{model}($o_{\textit{va}}, \theta$)
		\STATE $\pi(o_{\textit{va}})\gets$ Greedy search ($\bm{\beta}$, $A$)
		\IF{$\operatorname{mean}(\operatorname{L}(\pi(o_{\textit{va}})))<L_{tmp} $}
		\STATE $L_{tmp} \gets \operatorname{mean}(\operatorname{L}(\pi(o_{\textit{va}})))$
		\STATE $\operatorname{Save}$($\theta$)
		\ENDIF
		\ENDFOR
	\end{algorithmic}
	\label{alg1}
\end{algorithm}

When given a TSP instance $o$, \sysname{} decodes its corresponding output to obtain a TSP tour $\pi$, where the probability of selecting the subsequent node at each stage is determined as follows:
\begin{equation}\label{eq34}
P_{\theta}(a_t|s_t)=
\begin{cases}
\beta(\pi_1), & \text{if } t=0, \\
\gamma_{t+1}(\pi_{t+1}),  & \text{if } 0<t<n, \\
1, & \text{if } t=n,
\end{cases}
\end{equation}
where $\beta(\pi_1)$ and $\gamma_t(\pi_t)$ denote the probability of choosing the first node and the $t$th node in the TSP solution, respectively. Subsequently, the probability of the TSP tour $\pi$ is computed based on the chain rule as follows:
\begin{equation}\label{eq35}
	P_{\theta}(\pi|o)=\prod_{t=0}^nP_{\theta}(a_t|s_t).
\end{equation}
Furthermore, we adopt the central self-critical from \cite{Ma2020} to accelerate model convergence. Thus, the exact value of~(\ref{eq30}) is obtained as follows:
\begin{equation}\label{eq36}
\begin{aligned}
\nabla\mathcal{L}(\theta|s)=&\frac{1}{n_{bs}}\sum^{n_{bs}}[\operatorname{L}(\pi)-\operatorname{L}(b(o))+\omega)]\\& \cdot\nabla(\operatorname{log}\, \beta(\pi_1)
+\sum_{i=2}^n\operatorname{log}\,\gamma_i(\pi_i)),
\end{aligned}
\end{equation}
\begin{equation}\label{eq37}
\omega=\frac{1}{n_{bs}}\sum^{n_{bs}}[\operatorname{L}(\pi)-\operatorname{L}(b(o))], \omega\in\mathbb{R}^{1},
\end{equation}
where $n_{bs}$ denotes the value of batch size.

The overall training process of \sysname{} is outlined in Algorithm~\ref{alg1}. Our implementation of \sysname{} (in PyTorch) is accessible online\footnote{URL: \href{https://github.com/xybFight/\sysname}{https://github.com/xybFight/\sysname} }.


\subsection{Complexity Analysis of \sysname{}}
In this subsection, we present the overall time and space complexity of \sysname{}. For a given TSP instance with $n$ nodes, \sysname{} involves two main computational processes: the forward process and the decoding process. The forward process comprises the following three key operations: 1)~A linear layer with a time complexity of $\mathcal{O}(n^2)$ and a space complexity of $\mathcal{O}(n^2)$; 2)~GNN modules with a time complexity of $\mathcal{O}(n^2)$ and a space complexity of $\mathcal{O}(n^2)$; and 3) FC layers with a time complexity of $\mathcal{O}(n^2)$ and a space complexity of $\mathcal{O}(n^2)$. The decoding process has a time complexity of $\mathcal{O}(n^2)$ and a space complexity of $\mathcal{O}(n)$. 

Taken together, \sysname{} has a time complexity of $\mathcal{O}(n^2)$ and a space complexity of $\mathcal{O}(n^2)$. Compared to conventional TSP solvers, such as the Christofides algorithm with a time complexity of $\mathcal{O}(n^2\log(n))$, \sysname{} demonstrates potential benefits and scalability to larger instances. Additionally, due to highly parallelized GPU computing, \sysname{}'s inference time is further reduced, and it can solve multiple TSP instances simultaneously. This provides a considerable inference speed advantage over other heuristics algorithms, including unsupervised learning-based methods such as GSOA \cite{Faigl2018}, which also has a time complexity of $\mathcal{O}(n^2)$. See Section~\ref{section:Experiments} for the actual computational times taken by \sysname{}.

\subsection{Adoption of \sysname{} in Solving Other COPs Beyond TSP} \label{Other_co}
Benefiting from the decoding process of NAR4TSP, which imposes TSP constraints only on the model output, it can be easily extended to solve other COPs beyond TSP by imposing problem-specific constraints on the decoding process. In this work, we demonstrate NAR4TSP's application to solve the Manhattan TSPs and CVRPs. 

Comparing to the Euclidean TSP, the Manhattan TSP has more real-world applications, such as navigation tasks in urban environments. To extend NAR4TSP for solving Manhattan TSPs, we simply replace the Euclidean distances between nodes with the Manhattan distances, without making any other changes.

CVRP extends TSP by introducing a depot node, a capacity constraint $\kappa$, and demand requests of each node $D_i$ that are smaller than the capacity constraint. A feasible tour for CVRP consists of multiple sub-tours, each starting from the depot node and visiting a subset of nodes, with the total demand not exceeding the capacity. All nodes, except for the depot, must be visited exactly once. To extend NAR4TSP to solve CVRPs, we concatenate the node coordinate $x^{v}_i \in \mathbb{R}^{(n+1)\times d}$ and the node demand feature $D_i/\kappa$, as input nodes, i.e., $x^{v}_i \in \mathbb{R}^{(n+1)\times(d+1)}$. Because CVRP uses the depot node as the starting node by default, we remove the pointer $\beta$ from NAR4TSP but retain the other mechanisms. We impose CVRP-related constraints on the decoding process to obtain feasible solutions and then employ the enhanced RL learning strategy to train our model for solving CVRPs.

\begin{table*}[!t]
  \centering
    \caption{ \small Performance of \sysname{} with comparisons with benchmarking methods on TSP50 and TSP100 instances. Symbol $T$ denotes the number of iterations in the improvement process; symbol $B$* denotes the beam search with the shortest tour heuristic used in \cite{Joshi2019}; ``optimality gap'' indicates the relative difference in ``average length'' between each model and Concorde; ``S time'' indicates the inference time of solving a single TSP instance on average; and ``T time, $n_{bs}$'' indicates the inference time with batch size $n_{bs}$ on all the 10,000 testing TSP instances. Because of the end-to-end models' parallel processing nature, the reported value of ``T time" does not equal to the multiplication of the corresponding ``S time" and the total number of instances. Symbols ``2-opt" and ``AS" denote the post-hoc graph search techniques used in \cite{liu2024}.
    }
    \resizebox{2\columnwidth}{!}{
    \begin{tabular}{lccccccccccc}
    \toprule
    \multirow{3}[3]{*}{\textbf{category}} & \multirow{3}[3]{*}{\textbf{method}} & \multirow{3}[3]{*}{\textbf{type}} & \multirow{3}[3]{*}{${N_{\textit{paras}}}$} & \multicolumn{4}{c}{\textbf{TSP50}} & \multicolumn{4}{c}{\textbf{TSP100}} \\
\cmidrule{5-12}      & \multicolumn{1}{c}{} & \multicolumn{1}{c}{} & \multicolumn{1}{c}{} & \multicolumn{1}{c}{\textbf{\tabincell{c}{average\\length}$\downarrow$}} & \multicolumn{1}{c}{\textbf{\tabincell{c}{optimality\\gap}$\downarrow$}} & \textbf{\tabincell{c}{S time\\(sec)}$\downarrow$} & \textbf{\tabincell{c}{T time\\(sec), $n_{bs}$}$\downarrow$} & \multicolumn{1}{c}{\textbf{\tabincell{c}{average\\length}}$\downarrow$} & \multicolumn{1}{c}{\textbf{\tabincell{c}{optimality\\gap}}$\downarrow$} & \textbf{\tabincell{c}{S time\\(sec)}$\downarrow$} & \textbf{\tabincell{c}{T time\\(sec), $n_{bs}$}$\downarrow$} \\
\midrule
\multirow{6}{*}{\tabincell{l}{Conventional\\ algorithms}}
    & Concorde~(2007) \cite{Applegate2007} & exact solver & N.A. & 5.689 & 0.00\% & 0.035 & $3.63\times10^2$, 1 & 7.765 & 0.00\% & 0.165 & $1.45\times10^3$, 1 \\
    & LKH3~(2017) \cite{Helsgaun2017} & heuristic & N.A. & 5.689 & 0.00\% & 0.027 & $2.81\times10^2$, 1 & 7.766 & 0.01\% & 0.112 & $1.13\times10^3$, 1 \\
    & VNS & heuristic & N.A. & 5.960  & 4.76\% & 1.982 & $1.99\times10^4$, 1 & 8.327 & 7.24\% &  9.665 & $9.67\times10^4$, 1 \\
    
    & ACO & heuristic & N.A. & 5.981  & 5.13\% & 32.671 & $3.27\times10^5$, 1 & 8.614 & 10.93\% &  58.19 & $5.91\times10^5$, 1 \\

    & Nearest Insertion & heuristic & N.A. & 6.978  & 22.66\% & 0.020 & $2.04\times10^1$, 1 & 9.676 & 70.07\% & 0.044 & $4.36\times10^1$, 1 \\
   & Fastest Insertion & heuristic & N.A. & 5.999  & 5.47\% & 0.024 & $2.41\times10^1$, 1 & 8.347 & 7.50\% & 0.059 & $5.82\times10^1$, 1 \\
    \midrule
    \multirow{4}[2]{*}{\tabincell{l}{Neural improvement\\ algorithms}}
    & Wu \emph{et al.}, $T$=1000~(2022) \cite{Wu2022} & RL, heuristic & - & 5.740 &  0.89\% & - & - & 8.010 & 3.16\% & - & - \\
      & Wu \emph{et al.}, $T$=3000~(2022) \cite{Wu2022} & RL, heuristic & - & 5.710 &  0.36\% & - & - & 7.910 & 1.87\% & - & - \\
   & Costa \emph{et al.}, $T$=1000~(2021) \cite{Costa2021} & RL, heuristic & 1.34M & 5.735  & 0.81\% & 11.291 & $1.26\times10^3$, 500 & 7.851 & 1.10\% & 12.981 & $2.20\times10^3$, 128 \\
   & Costa \emph{et al.}, $T$=2000~(2021) \cite{Costa2021} & RL, heuristic & 1.34M & 5.718  & 0.52\% & 22.413 & $2.58\times10^3$, 500 & 7.820 & 0.71\% & 26.322 & $4.32\times10^3$, 128 \\
   \midrule
    \multirow{12}[2]{*}{\tabincell{l}{End-to-end models \\ with greedy search}} & Vinyals \emph{et al.}~(2015) \cite{Vinyals2015} & SL, AR & - & 7.660 & 34.64\% & - & - & - & - & - & - \\
      & Bello \emph{et al.}~(2017) \cite{Bello2017} & RL, AR & - & 5.950 & 4.59\% & - & - & 8.300  & 6.89\% & - & - \\
      & Khalil \emph{et al.}~(2017) \cite{Khalil2017} & RL, AR & - & 5.990 & 5.29\% & - & - & 8.310 & 7.02\% & - & - \\
      & Deudon \emph{et al.}~(2018) \cite{Deudon2018} & RL, AR & - & 5.920 & 4.06\% & - & - & 8.420 & 8.44\% & - & - \\
      & Kool \emph{et al.}~(2019) \cite{Kool2019} & RL, AR & 0.71M & 5.783 & 1.65\% & 0.079 & $4.68\times10^0$, 500 & 8.097 & 4.28\% & 0.154 & $1.88\times10^1$, 128\\
      & Bresson and Laurent~(2021) \cite{Bresson2021} & RL, AR & 1.41M & 5.707  & 0.32\% & 0.146 & $4.95\times10^0$, 500 & \textbf{7.876} & \textbf{1.43\%} & 0.277 & $2.24\times10^1$, 128 \\
      & Jung \emph{et al.}~(2023) \cite{Jung2023} & RL, AR & 1.43M & 5.754  & 1.13\% & 0.163 & $5.16\times10^0$, 500 & 7.985 & 2.83\% & 0.240 & $2.38\times10^1$, 128 \\
        & Yang \emph{et al.}~(2023) \cite{Yang2023} & RL, AR & 1.08M & 5.874  & 3.25\% & 0.126 & $\bm{3.96\times10^0}$, 500 & 8.175 & 5.29\% & 0.245 & $\bm{1.87\times10^1}$, 128 \\
      
       & Fellek \emph{et al.}~(2024) \cite{Fellek2024} & RL, AR & - & 5.747  & 0.86\% & - & - & 8.021 & 3.35\% & - & - \\
          & Liu \emph{et al.}~(2024) \cite{liu2024} + 2-opt +AS & SL, NAR & - & \textbf{5.699}  & \textbf{0.17\%} & - & - & 7.943 & 2.03\% & - & - \\
      & Joshi \emph{et al.}~(2019) \cite{Joshi2019} & SL, NAR & 11.05M & 5.870 & 3.18\% & 0.048 &  $8.02\times10^1$, 250 & 8.410 & 8.31\% & 0.090 & $3.70\times10^2$, 64 \\
      & \textbf{\sysname~(ours)} & RL, NAR & 0.91M & 5.752 & 1.11\% & \textbf{0.022} & $9.49\times10^0$, 500 & 7.899  & 1.74\% & \textbf{0.034} & $3.58\times10^1$, 128 \\
    \midrule
    \multirow{10}[2]{*}{\tabincell{l}{End-to-end models \\ with beam search}} & Kool \emph{et al.}, $B$=1280~(2019) \cite{Kool2019} & RL, AR & 0.71M & 5.716 & 0.47\% & 0.132 & $1.07\times10^3$, 1 & 7.939 & 2.24\% & 0.306 & $2.88\times10^3$, 1 \\
      & Kool \emph{et al.}, $B$=2560~(2019) \cite{Kool2019} & RL, AR & 0.71M & 5.713 & 0.42\% & 0.192 & $1.51\times10^3$, 1 & 7.931 & 2.14\% & 0.462 & $4.50\times10^3$, 1 \\
      & Bresson and Laurent, $B$=100~(2021) \cite{Bresson2021} & RL, AR & 1.41M & 5.692 & 0.05\% & 0.169 & $3.08\times10^2$, 150 & 7.818 & 0.68\% & 0.294 & $8.73\times10^2$, 80 \\
      & Bresson and Laurent, $B$=1000~(2021) \cite{Bresson2021} & RL, AR & 1.41M & \textbf{5.691} & \textbf{0.04\%} & 0.280 & $1.76\times10^3$, 15 & \textbf{7.803} & \textbf{0.49\%} & 0.781 & $6.48\times10^3$, 5 \\
      
      & Jung \emph{et al.}, $B$=1000~(2023) \cite{Jung2023} & RL, AR & 1.43M & 5.697 & 0.09\% & 0.241 & $1.16\times10^3$, 20 & 7.861 & 1.24\% & 0.732 & $3.88\times10^3$, 10 \\
        & Yang \emph{et al.}, $B$=1280~(2023) \cite{Yang2023} & RL, AR & 1.08M & 5.726 & 0.65\% & 0.310 & $2.62\times10^3$, 15 & 7.953 & 2.43\% & 0.874  & $6.31\times10^3$, 5 \\
      
      & Joshi \emph{et al.}, $B$=1280~(2019) \cite{Joshi2019} & SL, NAR & 11.05M & 5.710 & 0.36\% & 0.073 & $1.45\times10^2$, 250 & 7.920 & 2.00\% & 0.188 & $6.31\times10^2$, 64 \\
      & Joshi \emph{et al.}, $B$*=1280~(2019) \cite{Joshi2019} & SL, NAR & 11.05M & 5.700 & 0.19\% & 4.295 & $1.98\times10^3$, 250 & 7.880 & 1.48\% & 8.974 & $4.57\times10^3$, 64 \\
      & \textbf{\sysname, $B$=100~(ours)} & RL, NAR & 0.91M & 5.712 & 0.40\% & \textbf{0.038} & $\bm{1.05\times10^1}$, 500 & 7.845 & 1.02\% & \textbf{0.065} & $\bm{4.01\times10^1}$, 128 \\
      & \textbf{\sysname, $B$=1000~(ours)} & RL, NAR & 0.91M & 5.705 & 0.28\% & \textbf{0.047} & $\bm{1.86\times10^1}$, 500 & 7.827 & 0.80\% & \textbf{0.113} & $\bm{7.26\times10^1}$, 128 \\
    \bottomrule
    \end{tabular}}
  \label{table1}%
\end{table*}%

\section{Experimental Results}
\label{section:Experiments}
This section presents a set of comprehensive evaluations of \sysname. We start by providing an overview of the hyper-parameter settings of \sysname, followed by an assessment of the model's performance in terms of solution quality, inference ability, and generalization ability. Next, we illustrate the decoding process and the overall path planning of \sysname{} through visualizations. Then, we assess the effectiveness of extending \sysname{} to solve the Manhattan TSPs and CVRPs. Finally, we conduct ablation studies on the unique GNN architecture of \sysname{} and the enhanced RL strategy proposed in this paper.

\subsection{Hyper-parameter Configuration}\label{hyper}
For all experiments conducted in this paper, we consistently set the hyper-parameters of \sysname. Specifically, \sysname{} consists of $N_g=6$ GNN modules and $N_f=2$ FC layers, with a hidden dimension of $h=128$ for each layer and the multi-head attention number of $8$, following \cite{Bresson2021}. The trade-off parameter $\delta$ in~(\ref{eq4}) is set to $0.2$, following \cite{Velickovic2018}. To determine a node's neighbors, we consider the top $\frac{n}{5}$-nearest nodes, following \cite{Joshi2019}. The validation set size $n_v$ is set to 10,000. Each epoch consists of $n_s=2,500$ batches of $n_{bs}=64$ instances, with all training instances being generated randomly on the fly, following \cite{Kool2019, Bresson2021}. The number of training epochs varies based on the TSP size, with $n_e=1,000$ and $n_e=2,000$ epochs used for $n=50$ and $n=100$, respectively, following \cite{Bresson2021}. We use Adam \cite{Kingma2014} as the optimizer, with a constant learning rate of 1e--4 for simplification, despite that we recommend fine-tuning the learning rate~(e.g., learning rate decay) for faster convergence. All experiments were conducted on a computer equipped with an Intel(R) Core(TM) i5-11400F CPU and an NVIDIA RTX 3060Ti GPU.

\subsection{Performance of \sysname}
\label{section:evaluation}
We conduct comprehensive experiments to assess the performance of \sysname{} in three key aspects, namely solution quality, inference latency, and generalization ability.

\subsubsection{\textbf{Solution quality}}
We follow the convention and report the performance of \sysname{} for solving 10,000 two-dimensional TSP instances \cite{Kool2019}, where all test instances are randomly generated by sampling the node locations based on the unit uniform distribution. Table~\ref{table1} presents the performance of our model compared to the benchmarking methods on 10,000 test instances of TSP50 and TSP100. The methods compared in Table~\ref{table1} are categorized into four groups, namely conventional algorithms~(including exact solver and heuristics), neural improvement algorithms, end-to-end models with greedy search, and end-to-end models with beam search. To ensure a fair comparison on the inference time on all 10,000 test instances, we intend to maintain an equal batch size $n_{bs}$ for all NN-based models. But limited by the GPU memory~(8GB), four models \cite{Kool2019, Bresson2021, Jung2023, Yang2023} have to be run with smaller batch sizes when using beam search, and one \cite{Joshi2019} has to use a smaller batch size than other models due to its large number of learnable parameters. For Concorde and LKH3, we used the default settings. For VNS, we set the maximum number of neighborhood searches to 10. For ACO, we configured a population size of 20 and set the number of iterations to 1,000.


Table~\ref{table1} shows that models employing beam search outperform those using greedy search due to the wider search range it offers for potential solutions. Nonetheless, the utilization of beam search significantly impacts the performance of AR models \cite{Kool2019, Bresson2021, Jung2023, Yang2023}, particularly in terms of inference time and parallel processing capability~(i.e., the ability to simultaneously process $n_{bs}$ input instances). In contrast, NAR models, including \sysname, remain largely unaffected. Despite the one-shot nature of NAR decoding \cite{Joshi2022}, \sysname{} achieves competitive solution quality compared to all benchmarking NN-based models, with the exception of only one AR model \cite{Bresson2021}. This could be attributed to the fact that the model in \cite{Bresson2021} considers the contribution of all previously visited nodes when determining the next node for travel. However, such an approach undoubtedly increases the computational burden. Remarkably, compared to the best solution quality obtained by \cite{Bresson2021} with beam search width $B=1,000$, \sysname{} demonstrates significant advantages in terms of the inference time, with $\frac{1.76\times10^3}{1.86\times10^1}\approx 94$ times improvement on 10,000 TSP50 instances and $\frac{6.48\times10^3}{7.26\times10^1}\approx 89$ times improvement on 10,000 TSP100 instances, while only experiencing a slight reduction in solution quality~($1-\frac{5.691}{5.705}\approx0.24\%$ and $1-\frac{7.803}{7.827}\approx0.30\%$, respectively). Compared to another NAR model \cite{Joshi2019}, \sysname{} demonstrates notable advancements while utilizing a significantly smaller number of learnable parameters, i.e., $\frac{11.05}{0.91}\approx12.14$ times. Specifically, \sysname{} exhibits a remarkable $\frac{1.45\times10^2}{1.86\times10^1}\approx7.8$ times improvement in inference time and a slight yet noteworthy $\frac{5.710}{5.705}-1\approx0.08\%$ enhancement in solution quality when solving 10,000 TSP50 instances using beam search. Similarity, when solving 10,000 TSP100 instances, \sysname{} exhibits $\frac{6.31\times10^2}{7.26\times10^1}\approx8.7$ times improvement in inference time and $\frac{7.920}{7.827}-1\approx1.19\%$ improvement in solution quality. Furthermore, \sysname{} outperforms all the other NN-based methods in terms of the averaged inference time, highlighting its ability to produce high-quality solutions with low latency. 
\begin{figure*}[!t]
\centering
\subfigure[Solving TSP instances of different sizes using greedy search.]{
\includegraphics[width=0.98\columnwidth]{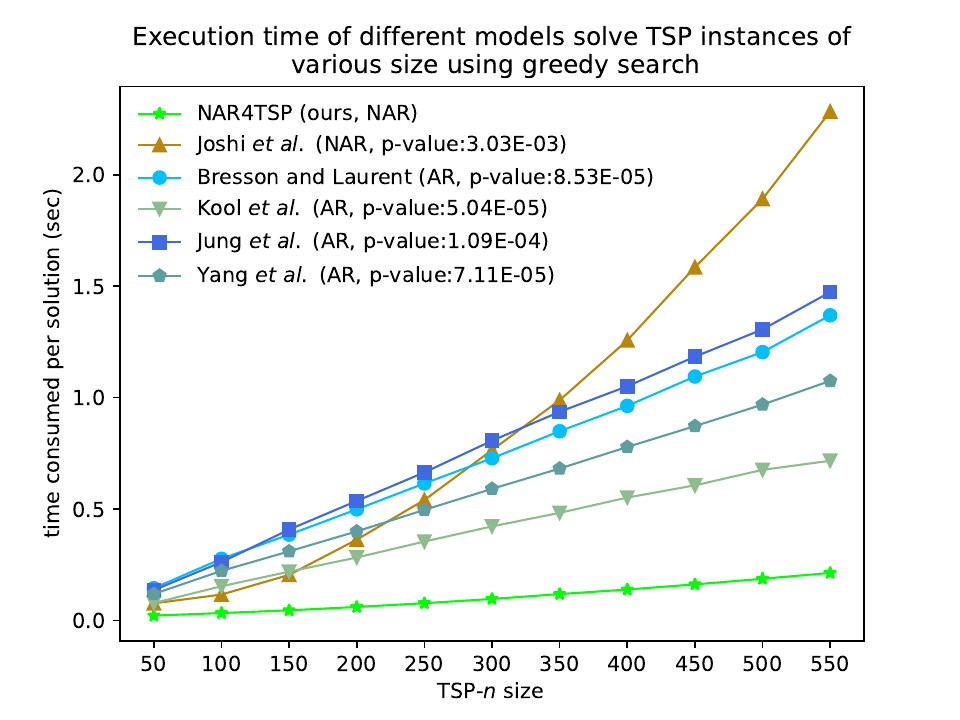}
\label{fig3a}}
\subfigure[Solving fixed-size TSP instances using beam search with different widths.]{
\includegraphics[width=0.98\columnwidth]{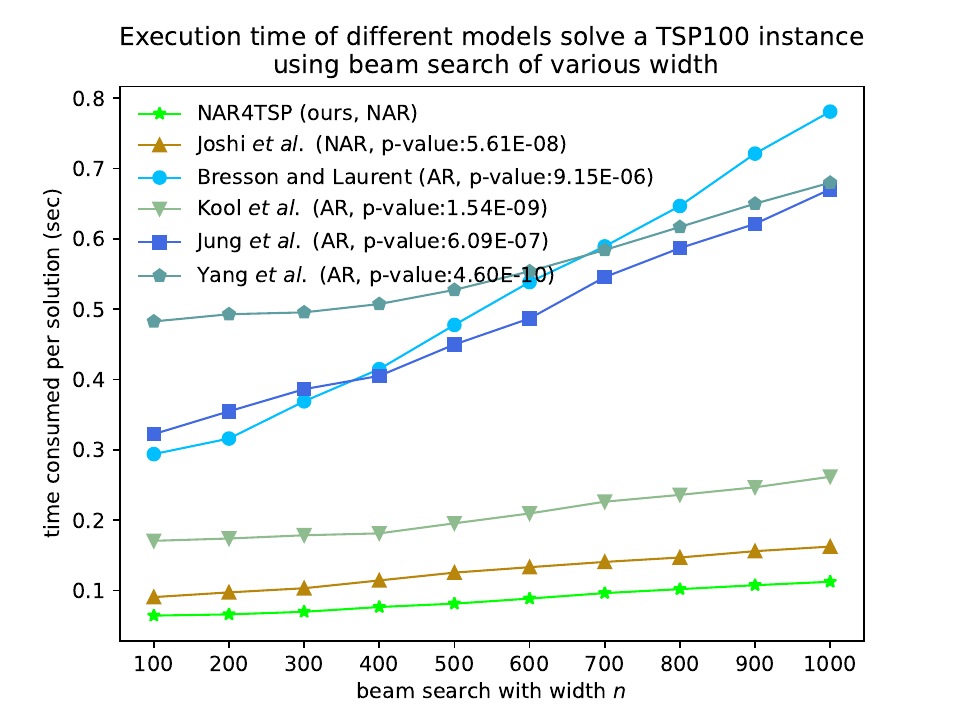}
\label{fig3b}}
\caption{Comparison on the inference time between our model and SOTA models with paired samples t-tests.\label{fig3}}
\end{figure*}

\subsubsection{\textbf{Inference latency}}
To comprehensively evaluate the inference speed of \sysname, we compare the time taken by \sysname{} and five SOTA models \cite{Joshi2019, Kool2019, Bresson2021, Jung2023, Yang2023} for solving TSP instances of varying sizes and beam widths. Figs.~\ref{fig3a} and~\ref{fig3b} illustrate the inference time taken by these models when using greedy search for TSP instances of different sizes and beam search with different widths for a fixed-size TSP instance, respectively. Furthermore, we conduct paired samples t-tests to determine whether significant differences exist between \sysname{} and the five SOTA models.

Fig.~\ref{fig3} illustrates that \sysname{} outperforms the other SOTA models in terms of inference speed. The disparity in inference time between \sysname{} and the other models becomes increasingly prominent as the number of nodes and beam width increase. Furthermore, the p-values obtained from other SOTA models are all below 0.05, leading to the conclusion that \sysname{} exhibits a significantly lower inference time than these models. These experimental results demonstrate that our \sysname{} model is much more applicable to real-time TSP tasks due to its low inference latency.

\begin{table*}[!t]
  \centering
  \caption{Generalization performance of applying models trained using TSP50 instances onto TSP$\{100, 200, 500, 1000\}$ instances}
  \resizebox{2\columnwidth}{!}{
    \begin{tabular}{lcccccccccc}
    \toprule
    \multirow{3}[2]{*}{\textbf{\tabincell{l}{model trained using\\TSP50 instances}}} & \multirow{3}[2]{*}{\textbf{type}} & \multirow{3}[2]{*}{\textbf{\tabincell{c}{beam\\width}}} & \multicolumn{2}{c}{\textbf{TSP100}} & \multicolumn{2}{c}{\textbf{TSP200}} & \multicolumn{2}{c}{\textbf{TSP500}} & \multicolumn{2}{c}{\textbf{TSP1000}} \\
          &       &       & {\textbf{\tabincell{c}{average\\length}}$\downarrow$} & \textbf{\tabincell{c}{optimality\\gap}$\downarrow$} & \multicolumn{1}{c}{\textbf{\tabincell{c}{average\\length}}$\downarrow$} & \textbf{\tabincell{c}{optimality\\gap}}$\downarrow$ & {\textbf{\tabincell{c}{average\\length}}$\downarrow$} & {\textbf{\tabincell{c}{optimality\\gap}}$\downarrow$} & {\textbf{\tabincell{c}{average\\length}}$\downarrow$} & {\textbf{\tabincell{c}{optimality\\gap}}$\downarrow$}\\
    \midrule
    \multirow{3}[2]{*}{Kool \emph{et al.}~(2019) \cite{Kool2019}} & \multirow{3}[2]{*}{RL, AR} & 1     & 8.144 & 4.88\% & \multicolumn{1}{c}{12.142} & 13.39\% &21.555 &30.39\% &33.756 &46.04\% \\
          &       & 1280  & 7.981 & 2.78\% & \multicolumn{1}{c}{12.904} & 20.49\% &29.583 &78.95\% &58.555 &153.33\% \\
          &       & 2560  & 7.966  & 2.59\% & \multicolumn{1}{c}{12.847} & 19.96\% &29.423 &77.99\% & - & - \\
    \midrule
    \multirow{3}[2]{*}{Bresson and Laurent~(2021) \cite{Bresson2021}} & \multirow{3}[2]{*}{RL, AR} & 1     & 8.023 & 4.11\% & \multicolumn{1}{c}{12.966} & 21.08\% &24.004 & 45.18\% &39.045 & 68.92\% \\
          &       & 100   & 7.929 & 3.12\% & \multicolumn{1}{c}{12.930} & 20.75\% &24.631 &48.99\% & 39.769 &72.05\%\\
          &       & 1000  & 7.906 & 1.82\% & \multicolumn{1}{c}{12.948 } & 20.91\% &24.691 &49.36\% &40.153 &73.71\%\\
        \midrule
    \multirow{2}[2]{*}{Jung \emph{et al.}~(2023) \cite{Jung2023}} & \multirow{2}[2]{*}{RL, AR} & 1  & 8.166  & 5.17\% & 12.435 & 16.13\% &22.611 & 36.78\% & 36.605 &58.37\%\\

          &       & 1000   & 7.990 & 2.90\% & 12.224  & 14.16\% &23.142 &39.99\% & 41.383 & 79.04\%\\
            \midrule
    \multirow{2}[2]{*}{Yang \emph{et al.}~(2023) \cite{Yang2023}} & \multirow{2}[2]{*}{RL, AR} & 1  & 8.291  & 6.77\% & 13.371 & 24.87\% & 25.082 &51.74\% & 41.887 &81.22\%\\
          &       & 1000   & 8.105 & 4.38\% & 12.865  & 20.14\% & 25.436 &53.87\% &42.848 &85.38\%\\
    \midrule
    \multirow{2}[2]{*}{Joshi \emph{et al.}~(2019) \cite{Joshi2019}} & \multirow{2}[2]{*}{RL, NAR} & 1     & 13.383 & 72.35\% & \multicolumn{1}{c}{21.688} & 102.54\% & - & -& -&- \\
          &       & 1280   & 12.222 & 57.40\% & \multicolumn{1}{c}{21.393} & 99.79\%& - & -& -&-  \\
    \midrule
%
    \multirow{3}[2]{*}{\textbf{\sysname~(ours)}} & \multirow{3}[2]{*}{RL, NAR} & 1 & 8.051 & 3.68\% & \multicolumn{1}{c}{11.828} & 10.45\% & 19.099 & 15.53\% & 27.479 & 18.88\%\\
          &       & 100 & 7.961 & 2.20\% & \multicolumn{1}{c}{11.617} & 8.49\%&18.864 &14.11\%  & 27.334 &18.26\%\\
          &       & 1000 & \textbf{7.898} & \textbf{1.72\%} & \textbf{11.532} & \textbf{7.69\%} & \textbf{18.740} & \textbf{13.36\%} & \textbf{27.245} & \textbf{17.87\%}\\
    \bottomrule
    \end{tabular}}%
  \label{table2}%
\end{table*}%

\begin{table*}[!t]
  \centering
  \caption{Generalization performance of applying models trained using TSP100 instances onto TSP$\{50, 200, 500, 1000\}$ instances}
  \resizebox{2\columnwidth}{!}{
    \begin{tabular}{lcccccccccc}
    \toprule
      \multirow{3}[2]{*}{\textbf{\tabincell{l}{model trained using\\TSP100 instances}}} & \multirow{3}[2]{*}{\textbf{type}} & \multirow{3}[2]{*}{\textbf{\tabincell{c}{beam\\width}}} & \multicolumn{2}{c}{\textbf{TSP50}} & \multicolumn{2}{c}{\textbf{TSP200}} & \multicolumn{2}{c}{\textbf{TSP500}}& \multicolumn{2}{c}{\textbf{TSP1000}} \\
          &       &       & {\textbf{\tabincell{c}{average\\length}}$\downarrow$} & \textbf{\tabincell{c}{optimality\\gap}$\downarrow$} & \multicolumn{1}{c}{\textbf{\tabincell{c}{average\\length}}$\downarrow$} & \textbf{\tabincell{c}{optimality\\gap}$\downarrow$} & {\textbf{\tabincell{c}{average\\length}}$\downarrow$} & \textbf{\tabincell{c}{optimality\\gap}$\downarrow$} & {\textbf{\tabincell{c}{average\\length}}$\downarrow$} & \textbf{\tabincell{c}{optimality\\gap}$\downarrow$}  \\
    \midrule
    \multirow{3}[2]{*}{Kool \emph{et al.}~(2019) \cite{Kool2019}} & \multirow{3}[2]{*}{RL, AR} & 1     & 5.935 & 4.32\% & \multicolumn{1}{c}{11.569} & 8.04\% &20.019 & 21.07\%& 31.183 & 34.91\%\\
          &       & 1280  & 5.821 & 2.34\% & \multicolumn{1}{c}{11.424} & 6.70\% & 22.576 & 36.57\% &42.629 & 84.42\%\\
          &       & 2560  & 5.816 & 2.23\% & \multicolumn{1}{c}{11.400} & 6.46\% & 22.508 & 36.15\%&- &-\\
    \midrule
    \multirow{3}[2]{*}{Bresson and Laurent~(2021) \cite{Bresson2021}} & \multirow{3}[2]{*}{RL, AR} & 1  & 6.214 & 9.22\% & \multicolumn{1}{c}{11.419} & 6.65\% &20.886 & 26.34\% & 33.364 &44.34\% \\
          &       & 100   & 5.860 & 3.00\% & \multicolumn{1}{c}{11.278 }  & 5.33\% & 20.737 & 25.45\% & 33.335 & 44.22\% \\
          &       & 1000  & 5.784 & 1.67\% & \multicolumn{1}{c}{11.240}  & 4.97\% & 20.699 & 25.21\% & 33.240 & 43.81\%  \\
    \midrule
    \multirow{2}[2]{*}{Jung \emph{et al.}~(2023) \cite{Jung2023}} & \multirow{2}[2]{*}{RL, AR} & 1  & 5.867 & 3.13\% & 11.576 & 8.10\% & 20.428 & 23.57\% & 31.853 & 37.81\%\\
          &       & 1000   & 5.751 & 1.08\% &  11.387 & 6.34\% & 20.410 &  23.46\% & 34.597 & 49.68\% \\
    \midrule
        \multirow{2}[2]{*}{Yang \emph{et al.}~(2023) \cite{Yang2023}} & \multirow{2}[2]{*}{RL, AR} & 1  & 5.910 & 3.88\% & 11.783 & 10.04\% &21.310 & 28.91\%& 35.334&52.87\% \\
          &       & 1000   & 5.793 & 1.83\% &  11.527 & 7.65\% &21.101 & 27.65\% & 35.598 &54.01\% \\
    \midrule
    \multirow{2}[2]{*}{Joshi \emph{et al.}~(2019) \cite{Joshi2019}} & \multirow{2}[2]{*}{RL, NAR} & 1  & 7.696 & 35.25\% & \multicolumn{1}{c}{16.827} & 57.14\% & -& -& -& -\\
          &       & 1280   & 7.422 & 30.44\% & \multicolumn{1}{c}{16.383 }  & 53.00\% & -& -& -& -\\
    \midrule

    \multirow{3}[2]{*}{\textbf{\sysname~(ours)}} & \multirow{3}[2]{*}{RL, NAR} & 1 & 5.808 & 2.09\% & \multicolumn{1}{c}{11.200} & 4.59\% & 18.767 & 13.52\% & 27.239  & 17.84\%\\
          &       & 100 & 5.739 & 0.88\% & \multicolumn{1}{c}{11.083} & 3.50\% &18.590 & 12.46\% &26.957 & 16.62\% \\
    &     & 1000 & \textbf{5.724} & \textbf{0.62\%} & \textbf{11.037}  & \textbf{3.07\%} & \textbf{18.520} & \textbf{12.03\%} &\textbf{26.824} & \textbf{16.05\%}\\
    \bottomrule
    \end{tabular}}%
  \label{table3}%
\end{table*}%

\subsubsection{\textbf{Generalization ability}}
The ability to generalize effectively across TSP instances in different sizes is a highly desirable characteristic for NN-based models to solve COPs \cite{Joshi2022}. Notably, \sysname{} shares parameters across all nodes, thus enabling it to remain robust to variations in the number of nodes. This property of \sysname{} allows the application of a pre-trained model to solve TSP instances of different sizes.


To assess the generalization ability of each model, we compare \sysname{} with five SOTA models \cite{Joshi2019, Kool2019, Bresson2021, Jung2023, Yang2023}. Table~\ref{table2} presents the performance of the models trained using TSP50 instances and applied to solve 10,000 TSP100, TSP200~(TSP with 200 nodes), TSP500~(TSP with 500 nodes), and TSP1000~(TSP with 1,000 nodes) instances. Similarly, Table~\ref{table3} demonstrates the performance of the models trained using TSP100 instances and applied to solve 10,000 TSP50, TSP200, TSP500, and TSP1000 instances. As the results presented in Tables~\ref{table2} and~\ref{table3}, \sysname{} and four AR models \cite{Kool2019, Bresson2021, Jung2023, Yang2023} perform well on TSP50, TSP100, and TSP200, while the other NAR model \cite{Joshi2019} does not. The plausible reason is overfitting that the model \cite{Joshi2019} uses an excessive number of parameters~(over 11 million) to memorize TSP training patterns, but the learned policy is still unable to generalize beyond the training sizes \cite{Joshi2022}. Importantly, \sysname{} exhibits superior generalization ability compared to all the other SOTA models, especially for solving large-scale problems (i.e., TSP500 and TSP1000). Furthermore, the results presented in Tables~\ref{table2} and~\ref{table3} indicate that the generalization performance of \sysname{} improves as the size of the TSP instances used for training increases, which is consistent with the commonsense that if the model is trained using complex examples, it should perform well on less complex ones.

\begin{table*}[!t]
\small
\centering
\caption{\footnotesize Zero-shot generalization performance of applying models trained using TSP50 instances onto 25 Euclidean and 10 non-Euclidean real-world TSPLIB instances. Optimality gaps indicate length differences compared to the optimal solutions from TSPLIB. Time indicates inference time.}\label{TSPLIB}
\  \resizebox{2\columnwidth}{!}{
    \begin{tabular}{ll|cccc|cccc|cccc|cccc}
    \toprule
    \multicolumn{2}{c|}{\multirow{3}[2]{*}{\textbf{instance}}} & \multicolumn{4}{c|}{\textbf{Bresson and Laurent~(2021) \cite{Bresson2021}}} & \multicolumn{4}{c|}{\textbf{Kool \emph{et al.}~(2019) \cite{Kool2019}}} &  \multicolumn{4}{c|}{\textbf{Jung~\emph{et al.}~(2023) \cite{Jung2023}}}  & \multicolumn{4}{c}{\textbf{\sysname~(ours)}} \\
    
    & & \multicolumn{2}{c}{\textbf{$B=1$}} & \multicolumn{2}{c|}{\textbf{$B=1000$}} & \multicolumn{2}{c}{\textbf{$B=1$}} & \multicolumn{2}{c|}{\textbf{$B=1000$}} & \multicolumn{2}{c}{\textbf{$B=1$}} & \multicolumn{2}{c|}{\textbf{$B=1000$}} & \multicolumn{2}{c}{\textbf{$B=1$}} & \multicolumn{2}{c}{\textbf{$B=1000$}} \\
    
    & \multicolumn{1}{c|}{} & \multicolumn{1}{c}{\textbf{\tabincell{c}{optimality\\gap}$\downarrow$}} & \multicolumn{1}{c}{\textbf{\tabincell{c}{time\\(sec)}}$\downarrow$} & \multicolumn{1}{c}{\textbf{\tabincell{c}{optimality\\gap}$\downarrow$}} &  \multicolumn{1}{c|}{\textbf{\tabincell{c}{time\\(sec)}}$\downarrow$} & \multicolumn{1}{c}{\textbf{\tabincell{c}{optimality\\gap}$\downarrow$}} & \multicolumn{1}{c}{\textbf{\tabincell{c}{time\\(sec)}}$\downarrow$} & \multicolumn{1}{c}{\textbf{\tabincell{c}{optimality\\gap}$\downarrow$}} & \multicolumn{1}{c|}{\textbf{\tabincell{c}{time\\(sec)}}$\downarrow$} & \multicolumn{1}{c}{\textbf{\tabincell{c}{optimality\\gap}$\downarrow$}} & \multicolumn{1}{c}{\textbf{\tabincell{c}{time\\(sec)}}$\downarrow$} & \multicolumn{1}{c}{\textbf{\tabincell{c}{optimality\\gap}$\downarrow$}} & \multicolumn{1}{c|}{\textbf{\tabincell{c}{time\\(sec)}}$\downarrow$} &
    \multicolumn{1}{c}{\textbf{\tabincell{c}{optimality\\gap}$\downarrow$}} & \multicolumn{1}{c}{\textbf{\tabincell{c}{time\\(sec)}}$\downarrow$} & \multicolumn{1}{c}{\textbf{\tabincell{c}{optimality\\gap}$\downarrow$}} & \multicolumn{1}{c}{\textbf{\tabincell{c}{time\\(sec)}}$\downarrow$} \\
    \midrule
    \multirow{28}[6]{*}{\rotatebox{90}{\textbf{Euclidean}}} & berlin52 & 93.56\% & 0.155  & 84.90\% & 0.283  & 222.32\% & 0.156  & 210.55\% & 0.188 & 267.37\% &0.153 & 297.98\% & 0.394 & 30.51\% & 0.029 & 10.94\% & 0.040  \\
    & ch130 & 190.07\% & 0.346  & 174.98\% & 1.271  & 608.59\% & 0.302  & 520.94\% & 0.572 & 623.57\%	& 0.326 & 562.14\% & 1.122 & 19.81\% & 0.057 & 12.97\% & 0.094  \\
    & ch150 & 187.71\% & 0.392  & 153.30\% & 1.681  & 765.45\% & 0.324  & 599.52\% & 0.746 & 720.82\%	& 0.439 & 655.12\%& 1.450   & 14.90\% & 0.062 & 8.89\% & 0.111 \\
     & d198 & 114.24\% & 0.527  & 111.65\% & 2.772  & 307.85\% & 0.480  & 884.74\% & 1.052 & 282.38\% & 0.573 & 995.16\% & 2.120  & 20.75\% & 0.086 & 15.67\% & 0.159 \\
    & d493 & 181.97\% & 1.259  & 172.41\% & 15.816  & 226.12\% & 1.193  & 1106.16\% & 5.432 & 450.63\% & 1.331 & 583.17\%	& 9.802  & 19.85\% & 0.287 & 14.98\% & 0.530 \\
    & eil101 & 104.13\% & 0.232  & 95.41\% & 0.804  & 372.40\% & 0.233  & 345.98\% & 0.417 & 326.55\%	& 0.300 & 407.15\%	& 0.799  & 15.62\% & 0.055 & 11.58\% & 0.088 \\ & eil51 & 80.84\% & 0.118  & 64.51\% & 0.269  & 268.03\% & 0.141  & 217.18\% & 0.185 & 255.87\% & 0.155 & 242.72\% & 0.369  & 14.10\% & 0.033 & 3.72\% & 0.044 \\    & eil76 & 95.76\% & 0.184  & 87.36\% & 0.516  & 373.93\% & 0.183  & 269.14\% & 0.271 & 337.17\% & 0.230 & 345.17\% & 0.578  & 14.34\% & 0.045 & 9.83\% & 0.075 \\
    & gil262 & 539.90\% & 0.674  & 506.75\% & 4.600  & 1022.72\% & 0.606  & 901.04\% & 1.858 & 939.61\% & 0.743 & 927.63\% & 3.333  & 22.96\% & 0.123 & 17.52\% & 0.236 \\
    & kroA100 & 171.14\% & 0.273  & 156.50\% & 0.792  & 730.25\% & 0.219  & 516.93\% & 0.374 & 755.84\% & 0.296 & 751.52\% & 0.915  & 39.01\% & 0.047 & 16.16\% & 0.095  \\
    & kroB150 & 186.67\% & 0.380  & 178.55\% & 1.655  & 713.21\% & 0.380  & 550.89\% & 0.795 & 941.20\% & 0.411 & 913.69\% & 1.400 & 20.17\% & 0.066 & 16.06\% & 0.109  \\
    & kroD100 & 152.91\% & 0.259  & 143.06\% & 0.801  & 517.61\% & 0.251  & 397.31\% & 0.404 & 698.75\% & 0.274 & 583.41\% & 0.779 & 17.81\% & 0.061 & 11.30\% & 0.073 \\
    & kroE100 & 160.57\% & 0.307  & 138.24\% & 0.810  & 704.46\% & 0.295  & 445.74\% & 0.400 & 770.32\% & 0.278 & 656.38\% & 0.820 & 19.06\% & 0.056 & 10.70\% & 0.074 \\
    & lin105 & 207.39\% & 0.287  & 187.51\% & 0.868  & 649.11\% & 0.284  & 424.61\% & 0.431 & 225.33\% & 0.293 & 684.89\% & 0.851 & 12.90\% & 0.048 & 8.23\% & 0.082  \\
    & lin318 & 301.79\% & 0.860  & 336.47\% & 6.848  & 538.74\% & 0.789  & 1110.76\% & 2.442 & 314.61\% & 0.862 & 661.62\% & 4.678 & 19.38\% & 0.158 & 15.33\% & 0.271 \\
    & pcb442 & 308.81\% & 1.192  & 292.29\% & 12.484  & 910.06\% & 1.082  & 1186.35\% & 4.380 & 509.00\% & 1.244 & 863.51\% & 7.968 & 23.28\% & 0.249 & 16.24\% & 0.421 \\
    & pr107 & 35.90\% & 0.280  & 28.14\% & 0.884  & 554.30\% & 0.374  & 670.63\% & 0.404 & 41.64\% & 0.307 & 606.72\% & 0.908 & 8.74\% & 0.045& 6.49\% & 0.088  \\
    & pr144 & 86.17\% & 0.382  & 66.82\% & 1.551  & 428.35\% & 0.349  & 889.62\% & 0.613 & 612.85\% & 0.414 & 999.26\% & 1.324 & 7.34\% & 0.061 & 6.86\% & 0.104 \\
    & pr226 & 141.45\% & 0.639  & 157.34\% & 3.594  & 80.08\% & 0.516  & 1435.42\% & 1.394 & 862.92\%	& 0.647 & 1524.86\% & 2.630 & 29.05\% & 0.102 & 8.45\% & 0.187 \\
    & pr264 & 118.66\% & 0.688  & 111.44\% & 4.875  & 314.77\% & 0.614  & 1663.57\% & 1.780 & 1007.34\% & 0.750 & 1832.27\% & 3.332 & 16.75\% & 0.124 & 10.58\% & 0.210 \\
    & pr76 & 136.02\% & 0.208  & 117.72\% & 0.483  & 221.80\% & 0.209  & 307.85\% & 0.262 & 168.01\% & 0.232 & 335.40\% & 0.578 & 17.15\% & 0.038 & 5.05\% & 0.058 \\
    & rat195 & 212.04\% & 0.503  & 219.91\% & 2.600  & 146.81\% & 0.480  & 727.44\% & 1.029 & 73.48\%	& 0.568 & 83.81\%	& 2.062 & 18.43\% & 0.118 & 12.55\% & 0.152 \\
    & rat575 & 337.21\% & 1.511  & 346.37\% & 21.855  & 172.43\% & 1.366  & 1401.64\% & 7.223 & 158.16\% & 1.652 & 244.69\% & 12.642 & 24.76\% & 0.396 & 17.57\% & 0.627  \\
    & rat99 & 161.02\% & 0.246  & 153.66\% & 0.770  & 136.68\% & 0.246  & 435.97\% & 0.391 & 120.64\%	& 0.269 & 229.15\% & 0.788 & 17.53\% & 0.046 & 10.95\% & 0.074 \\
    & st70 & 108.56\% & 0.204  & 91.82\% & 0.426  & 412.01\% & 0.176  & 319.66\% & 0.241 & 413.78\% & 0.181 & 398.52\% & 0.495 & 17.41\% & 0.033 & 9.14\% & 0.053 \\
    \cmidrule{2-18}
    \multicolumn{2}{c}{\textbf{average}} & \textbf{176.58\%} & \textbf{0.484}  & \textbf{167.08\%} & \textbf{3.573}  & \textbf{455.92\%} & \textbf{0.450}  & \textbf{701.59\%} & \textbf{1.331}  & \textbf{475.11\%} & \textbf{0.517} & \textbf{655.44\%} & \textbf{2.485} & \textbf{19.26\%} & \textbf{0.097} & \textbf{11.51\%} & \textbf{0.162}  \\
    \cmidrule{2-18}
        
    \multicolumn{18}{c}{\textbf{Normalized 25 Euclidean instances}}\\
    \cmidrule{2-18}
    \multicolumn{2}{c}{\textbf{average}} & \textbf{24.05\%} & \textbf{0.479}  & \textbf{25.81\%} & \textbf{3.461}  & \textbf{17.46\%} & \textbf{0.462}  & \textbf{26.51\%} & \textbf{1.322}  & \textbf{15.26\%} & \textbf{0.531} & \textbf{15.02\%} & \textbf{2.662} & \textbf{12.26\%} & \textbf{0.102} & \textbf{8.65\%} & \textbf{0.169}  \\
    \hline
    \hline
    \multirow{11}[5]{*}{\rotatebox{90}{\textbf{non-Euclidean}}} & ali535 & 1574.71\%  & 1.365  & 1457.62\% & 18.772 & 1558.06\% & 1.298  & 1475.36\% & 6.180 & 1612.46\% & 1.470 & 1515.20\% & 10.619 & 26.45\% & 0.269 & 26.45\% & 0.598 \\
    & att48 & 389.66\%& 0.139  & 279.67\% & 0.269 & 339.34\%  &   0.112   & 172.19\% & 0.176 & 290.08\% & 0.145 & 308.09\% & 0.320 & 38.94\% & 0.026 & 19.35\% & 0.042 \\
    & att532 & 1259.29\%  & 1.371  & 1649.22\% & 18.320 & 1019.60\% & 1.301  & 1354.58\% & 6.165 & 1138.85\% & 1.471 & 1290.03\% & 10.598 & 30.36\% & 0.266 & 24.06\% & 0.591 \\
    & bayg29 & 186.02\% & 0.101  & 113.98\% & 0.173 & 188.39\% &  0.087 & 115.65\% & 0.124 & 195.03\% & 0.111 & 128.39\% & 0.171 & 32.30\% & 0.021 & 1.37\% & 0.031 \\
    & bays29 & 181.39\% & 0.101  & 118.81\% & 0.181 & 191.04\% &  0.079 & 116.24\% & 0.124 & 197.28\% & 0.105 & 133.91\% & 0.169 & 3.71\% & 0.022 & 3.61\% & 0.029 \\
    & burma14 & 93.47\% & 0.095  & 63.86\% & 0.130 & 37.29\% &  0.044 & 39.06\% &   0.119 & 37.29\% & 0.066 & 87.69\% & 0.129 & 22.90\% & 0.018 & 6.53\% & 0.022 \\
    & dantzig42 & 123.46\% & 0.138 & 252.07\% & 0.215 & 22.32\% &  0.100  & 181.69\% & 0.151 & 168.24\% & 0.128 & 334.05\% & 0.192 & 7.44\% & 0.026 & 7.30\% & 0.041 \\
    & pa561 & 967.57\% & 1.431  & 1199.38\% & 20.661 & 111.51\% & 1.355  & 1149.40\% & 6.700 & 592.22\% & 1.601 & 746.87\% & 12.724 & 22.98\% & 0.321 & 18.93\% & 0.613 \\
    & ulysses16 & 91.59\% & 0.071  & 51.51\% & 0.124 & 53.13\% &  0.045 & 37.66\% & 0.110 & 75.68\% & 0.065 & 58.89\% & 0.130 & 25.53\% & 0.018 & 1.33\% & 0.020 \\
    & ulysses22 & 84.53\% &  0.085 & 93.74\% & 0.149 & 131.44\% &   0.061& 76.99\% & 0.119 & 91.19\% & 0.078 & 110.98\% & 0.158 & 19.82\% & 0.019 & 1.90\% & 0.026 \\
    \cmidrule{2-18}
    \multicolumn{2}{c}{\textbf{average}} & \textbf{495.17\%} & \textbf{0.490}  & \textbf{527.99\%} & \textbf{5.897}  & \textbf{365.21\%} & \textbf{0.448}  & \textbf{471.88} & \textbf{1.995}  & \textbf{439.83} & \textbf{0.524} & \textbf{471.41\%} & \textbf{3.520} & \textbf{23.04\%} & \textbf{0.101} & \textbf{11.08\%} & \textbf{0.201}  \\
    \bottomrule
\end{tabular}}%

\end{table*}

\begin{table*}[!t]
  \centering
  \caption{Performance of \sysname{} with comparisons with benchmarking methods on 25 real-world Euclidean TSPLIB instances}
  \resizebox{2\columnwidth}{!}{
    \begin{tabular}{l|cccc|ccccc}
    \toprule
   \multirow{2}[2]{*}{\textbf{Average}} & \multicolumn{4}{c|}{\textbf{Conventional algorithms}} &  \multicolumn{5}{c}{\textbf{End-to-end models$^{\dag}$}}\\
    & Concorde & LKH3& VNS & ACO & Bresson and Laurent~(2021) \cite{Bresson2021} & Kool \emph{et al.}~(2019) \cite{Kool2019} & Jung \emph{et al.} (2023) \cite{Jung2023} & Yang \emph{et al.} (2023) \cite{Yang2023} & NAR4TSP (ours) \\
    \midrule
    \textbf{optimality gap $\downarrow$} &\textbf{0.00\%} &\textbf{0.00\%} &7.74\% & 20.26\%& 8.93\% & 18.21\% & 11.71\%&14.21\% &6.21\%\\
    \textbf{time (sec)} $\downarrow$ &3.821 &   2.490& 91.358&101.24 & 3.450 & 1.345& 2.662& 2.142& \textbf{0.172}\\
    \bottomrule
    \end{tabular}}%
  \label{table_lib}%

  \begin{tablenotes}
    \footnotesize
    \item[2] $\dag$: We employ the baseline models trained using TSP100 instances with beam ($B=1000$) search and perform normalization for better solution quality.
    \end{tablenotes}
\end{table*}%

Nonetheless, in Tables~\ref{table2} and~\ref{table3}, the node coordinates in the test instances are all sampled based on the unit uniform distribution, which is consistent with the training data. We further evaluate the generalization ability of \sysname{} and three SOTA AR models \cite{Kool2019, Bresson2021, Jung2023} using the widely adopted TSPLIB\footnote{URL: \href{http://comopt.ifi.uni-heidelberg.de/software/TSPLIB95/}{http://comopt.ifi.uni-heidelberg.de/software/TSPLIB95/}} dataset. TSPLIB is a repository of TSP instances widely used to assess the effectiveness of algorithms for solving TSPs. To this end, we randomly select 25 real-world Euclidean TSP instances with sizes ranging from 51 to 575 nodes and 10 non-Euclidean TSP instances with sizes ranging from 14 to 561 nodes from the TSPLIB dataset as test instances. Unlike the training instances where the node coordinates are bounded between 0 and 1, the node coordinates of the real-world instances in TSPLIB are integers that span from one to thousands and in some cases, even negative values. Although the randomly sampled training instances differ significantly from the TSPLIB instances, particularly some of them are motivated by the printed circuit board assembly, the objective remains the same: to produce a Hamiltonian cycle that minimizes the sum of edge weights in the path. The end-to-end NN-based model learns a mapping from model inputs to outputs, i.e., from a TSP graph to a Hamiltonian cycle with the shortest path (ideally). Leveraging this property, models trained on random instances can be reasonably applied to solve TSPLIB instances as shown in prior studies \cite{Wang2024, Wu2022}. To maximally simulate the unknown real-world scenarios, we refrain from applying any preprocessing methods, such as normalization, rotation, or transformation, to the test instances. Instead, we straightforwardly apply the models trained using the TSP50 instances to solve the 35 real-world instances. This setting is widely known as zero-shot generalization \cite{Joshi2022}. A model capable of generating good solutions in such cases signifies its ability to recognize and learn the underlying patterns and insights of TSPs, making it suitable for diverse applications, as opposed to superficial pattern recognition \cite{Joshi2022}.
\begin{figure*}[!t]
\centering
\includegraphics[width=1.9\columnwidth]{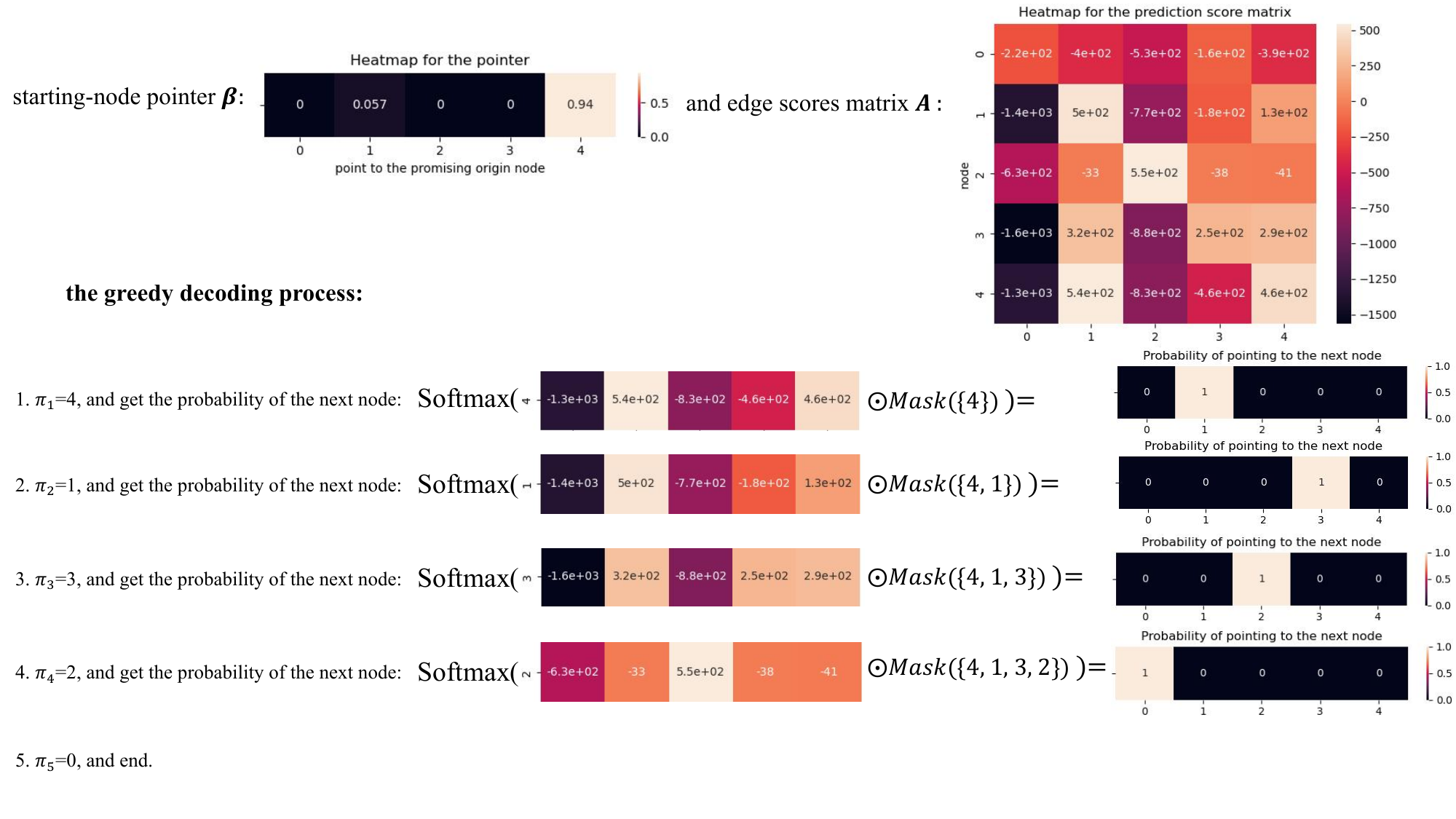}
\caption{Visualization of a TSP5 instance's greedy decoding process.}
\label{fig4}
\end{figure*}
\begin{figure*}[!t]
\centering
\subfigure[TSP50 instances.]{
\includegraphics[width=0.985\columnwidth]{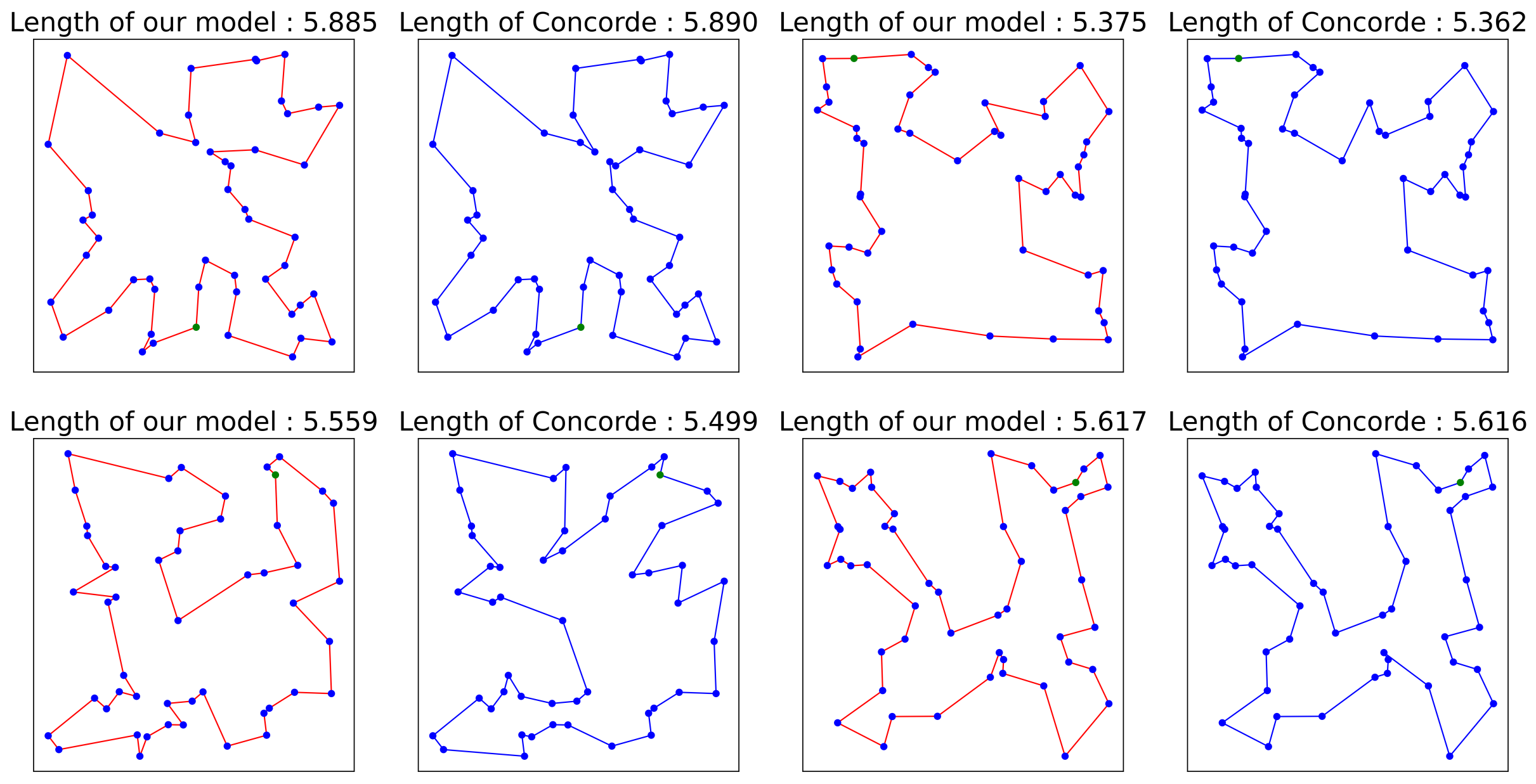}
\label{fig5a}}
\subfigure[TSP100 instances.]{
\includegraphics[width=0.985\columnwidth]{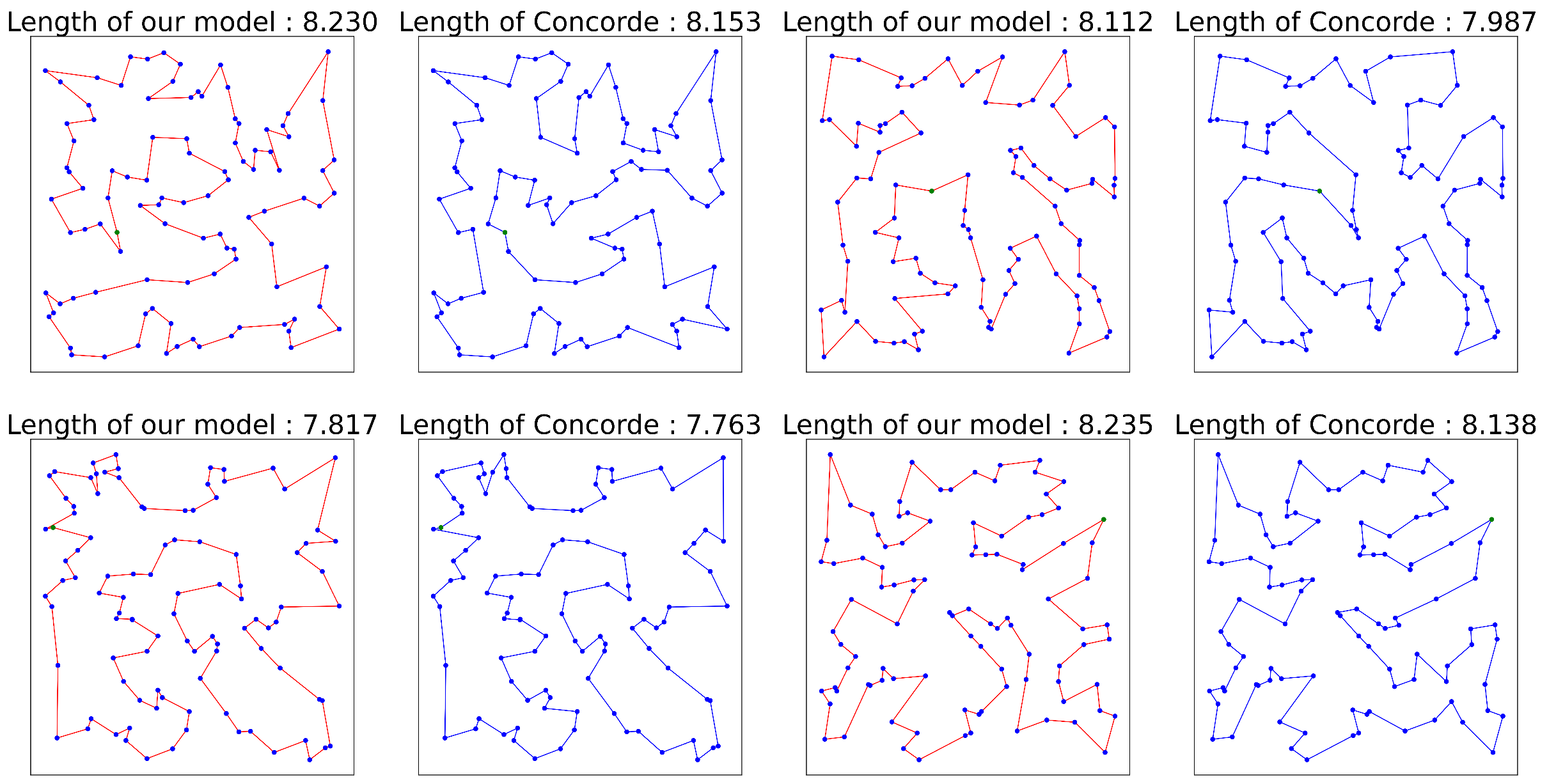}
\label{fig5b}}
\caption{Visualization of solutions produced by \sysname{} using greedy search and Concorde, respectively.\label{fig5}}
\end{figure*}

As shown in Table~\ref{TSPLIB}, the three SOTA models \cite{Kool2019, Bresson2021, Jung2023} exhibit unsatisfactory results. Although increasing the beam width $B$ generally improves the solution quality by preserving the top-$B$ candidate solutions with the highest probability, the model \cite{Bresson2021} shows a $1-\frac{167.08\%}{176.58\%}\approx5.37\%$ improvement in performance for Euclidean instances and a $\frac{572.99\%}{495.17\%}-1\approx15.71\%$ decline for non-Euclidean instances as $B$ increases from $1$ to $1000$. On the other hand, the performance of the other two models \cite{Kool2019, Jung2023} even decreases by $\frac{701.59\%}{455.92\%}-1\approx53.88\%$ and $\frac{655.44\%}{475.11\%}-1\approx37.95\%$ for Euclidean instances and by $\frac{471.88\%}{365.21\%}-1\approx29.21\%$ and $\frac{471.41\%}{439.83\%}-1\approx7.18\%$ for non-Euclidean instances, respectively. This finding indicates that these models have difficulties in handling TSP cases formulated differently from the training instances. In contrast, \sysname{} outperforms these models by showcasing solution quality improvements of $1-\frac{11.51\%}{19.26\%}\approx40.25\%$ and $1-\frac{11.08\%}{23.04\%}\approx51.91\%$ for Euclidean and non-Euclidean instances, respectively, when the beam width $B$ increases from $1$ to $1000$. Furthermore, \sysname{} exhibits significantly lower computation time, accounting for only $\frac{0.162 + 0.201}{3.573 + 5.897}\approx3.83\%$ of the total average time taken by \cite{Bresson2021}, $\frac{0.162 + 0.201}{1.331 + 1.995}\approx10.91\%$ of the total average time taken by \cite{Kool2019}, and $\frac{0.162+0.210}{2.485 + 3.520}\approx6.05\%$ of the total average time taken by \cite{Jung2023}. This finding highlights the remarkable time efficiency of our \sysname{} model.

In addition, we preprocess the 25 Euclidean TSPLIB instances by normalizing their node coordinates and subsequently compare the performance of \sysname{} with the three SOTA models on the normalized instances. It is worth noting that we do not normalize the non-Euclidean TSPLIB instances, because normalization may alter the relative position of nodes in non-Euclidean TSP, unlike in Euclidean TSP. As indicated by the results presented in Table~\ref{TSPLIB}, \sysname{} again outperforms the three SOTA models. Remarkably, \sysname{} achieves better performance even without requiring instances normalization (optimality gap = 11.51\% when $B$ = 1,000) comparing to the three SOTA models with instances normalization, despite the significant improvement in solution quality attained by the latter after normalization. Furthermore, given that models trained on larger-scale TSP instances exhibit better generalization performance (see Tables~\ref{table2} and~\ref{table3}), we report the results of end-to-end models trained on TSP100s and applied to solve the normalized 25 Euclidean TSPLIB instances, along with those of four conventional TSP solvers. As shown in Table~\ref{table_lib}, \sysname{} outperforms the end-to-end models in terms of solution quality and has significantly lower inference times.


\subsection{Visualization of the Decoding Process and the Overall Path Planning of \sysname}
\label{section:Visualization}
We employ two visualization methods to demonstrate the feasibility of implementing \sysname{} in an end-to-end manner and the effectiveness of \sysname, respectively.

Firstly, we visualize the decoding process of \sysname{} using a randomly generated TSP5~(TSP with five nodes) instance as input. We depict the probability distribution of selecting the next node at each decoding stage. As shown in Fig.~\ref{fig4}, our model produces a deterministic sequence by decoding a pointer and a matrix of edge scores, indicating its ability to learn an end-to-end mapping from the TSP information to the solution. Additionally, the heatmap of the score matrix $A$ reveals that \sysname{} goes beyond learning simple domain relations, such as the distance relationship between nodes. In this case, the diagonal elements are expected to consistently exhibit the largest values within their respective rows, because they represent the proximity of a node to itself, which should yield the highest similarity score among all nodes. However, it is worth noting that matrix $A$ contains numerous elements with values surpassing those of the diagonal elements in their corresponding rows. This observation implies the existence of nodes that exhibit stronger connections to other nodes than to themselves, showcasing that our model has learned the intrinsic mapping relationships correctly.

Secondly, we visualize the overall path planning of \sysname{} by randomly selecting a set of TSP50 and TSP100 instances, respectively and comparing the solutions produced by \sysname{} with greedy search and the optimal solution given by Concorde. As shown in Fig.~\ref{fig5}, \sysname{} yields on-par tour lengths comparing with those of Concorde, thus demonstrating the effectiveness of our model in solving TSPs.


\subsection{Performance of \sysname{} on Solving Other COPs}

In this subsection, we evaluate the effectiveness of extending NAR4TSP to solve other COPs. Specifically, we extend \sysname{} to solve the Manhattan TSP50 and CVRP50 (CVRP with 50 nodes) instances, respectively~(see problem definitions in Section~\ref{Other_co}). Both sets of experiments are conducted in the same environment used for solving TSPs (see Section~\ref{hyper}).  For the Manhattan TSP50, the training and testing instances are randomly generated by sampling the node locations based on the unit uniform distribution. For CVRP50, we sample the coordinates of nodes (including the depot) of training and testing instances uniformly at random from the unit interval. The demand $D_i$ for each node is defined as $\frac{\delta_i}{40}$, where $\delta_i$ denotes a uniformly sampled discrete value from $\{1, 2, \dots, 9\}$. The capacity $\kappa$ is set to $1$. 

\begin{table*}[!t]
  \centering
  \caption{\footnotesize Performance of \sysname{} on solving Manhattan TSPs and CVRPs. ``optimality gap'' indicates the relative difference in ``average length'' between our model and LKH3. }\label{oth_CO}
  \resizebox{2\columnwidth}{!}{
    \begin{tabular}{l|cccc|cccc}
    \toprule
    \multirow{3}{*}{\textbf{Method}}& \multicolumn{4}{c}{\textbf{Manhattan TSP50}} & \multicolumn{4}{c}{\textbf{CVRP50}}\\
    & \textbf{\tabincell{c}{average\\length}$\downarrow$} & \textbf{\tabincell{c}{optimality\\gap}$\downarrow$} & \textbf{\tabincell{c}{S time\\(sec)}$\downarrow$} & \textbf{\tabincell{c}{T time\\(sec), $n_{bs}$}$\downarrow$}&\textbf{\tabincell{c}{average\\length}$\downarrow$} & \textbf{\tabincell{c}{optimality\\gap}$\downarrow$} & \textbf{\tabincell{c}{S time\\(sec)}$\downarrow$} & \textbf{\tabincell{c}{T time\\(sec), $n_{bs}$}$\downarrow$}\\
    \midrule
    LKH3 & 6.985 & 0.00\% & 0.029 & $2.76\times10^2$, 1 & 10.360 & 0.00\% & 5.280& $5.40\times10^3$, 1\\
    NAR4TSP (greedy) & 7.197 & 3.03\% &0.023 & $9.72\times10^0$, 500 & 11.321& 9.27\%&  0.046&$1.02\times10^1$, 400\\
    NAR4TSP ($B=1000$)  &7.110 & 1.79\% & 0.052& $1.83\times10^1$, 500& 10.870& 4.92\%& 0.073&$2.18\times10^1$, 400\\
    \bottomrule
    \end{tabular}}
    
\end{table*}%

As shown in Table~\ref{oth_CO}, \sysname{} exhibits excellent performance on both the Manhattan TSPs and CVRPs, providing comparable solution quality and notably reduced inference times. These results demonstrate the effectiveness of extending NAR4TSP to solve other COPs.

\begin{table}[!t]
  \centering
  \caption{Performance comparison of the GCN of \cite{Joshi2019} and \sysname}.
  \resizebox{1\columnwidth}{!}{
    \begin{tabular}{|cccccc|}
    \toprule
    \multicolumn{2}{|c}{\textbf{model architecture}} & \multicolumn{1}{c}{${N_{\textit{paras}}}$} & \multicolumn{1}{c}{\textbf{\tabincell{c}{average\\length}$\downarrow$}} & \multicolumn{1}{c}{\textbf{\tabincell{c}{S time\\(sec)}$\downarrow$}} & \multicolumn{1}{c|}{\textbf{\tabincell{c}{T time\\(sec), $n_{bs}$}$\downarrow$}} \\
    \midrule
    \multicolumn{2}{|c}{GCN used in \cite{Joshi2019}} &11.05M &6.398 &0.046 &79.21, 250 \\
    \multicolumn{2}{|c}{\textbf{GNN (ours)}} &0.91M &5.752 &0.021 &9.39, 500 \\
    \multicolumn{2}{|c}{GNN without the pointer mechanism} &0.91M & 6.117 & \textbf{0.020} & \textbf{9.21}, 500 \\
    
    \multirow{3}{*}{GNN with all nodes as the starting node} & min & \multirow{3}{*}{0.91M} & \textbf{5.748} & \multirow{3}{*}{0.029} & \multirow{3}{*}{19.14, 500} \\

    & max & & 7.466 & &  \\
    
    & mean & & 6.671 & & \\
    \bottomrule
    \end{tabular}}
    \label{table5}
\end{table}%

\subsection{Effectiveness of the Novel GNN Architecture and the Enhanced RL Strategy}
\label{section:effectiveness}
In this subsection, we evaluate the effectiveness of the novel GNN architecture and the enhanced RL strategy proposed in this paper. Specifically, we use 10,000 TSP50 instances as the test set and systematically evaluate the performance by varying the components under investigation. Specifically, we conduct three ablation studies to evaluate the effectiveness of the GNN architecture. Firstly, we replace the proposed GNN architecture with the GCN introduced in \cite{Joshi2019}. Secondly, we remove the pointer~\bm{$\beta$} from \sysname{} and adopt the convention of \cite{Nowak2018, Joshi2019} to use the first node as the starting position for both training and testing. Thirdly, we employ an inference method akin to the one used in \cite{Kwon2020}, selecting all nodes as the potential starting position for inference, which inevitably increases the computational cost and inference latency.

We present the ablation study results in Table~\ref{table5}. Firstly, although the architecture of \cite{Joshi2019} employs more learnable parameters than \sysname, its average solution length on TSP50 is merely 6.398~(using greedy search). This represents a $\frac{6.398}{5.752}-1\approx11.23\%$ decrease compared to \sysname{} using our proposed GNN architecture, thus highlighting the effectiveness of the novel GNN architecture. Secondly, the absence of the pointer mechanism in \sysname{} leads to a relative  performance drop of $\frac{6.117}{5.752}-1\approx6.34\%$ on the test set, underscoring the contribution of the learnable pointer. It is worth noting that even in the absence of the pointer mechanism, \sysname{} achieves an average solution length that is on-par with that of the SOTA NAR model \cite{Joshi2019} along with a $\frac{8.02\times10^1}{9.21\times10^0}\approx8.7$-fold speed enhancement (see Table~\ref{table1}). This result can be attributed to the well-designed GNN architecture that facilitates TSP representation extraction (see Section~\ref{section:architecture}). Thirdly, when each node is considered as the starting position, the minimum solution length is only $1-\frac{5.748}{5.752}\approx0.06\%$ better than the solution length derived based on the \sysname{}'s determined pointer $\bm{\beta}$. It is worth noting that the former must be less than or equal to the latter. Equality is only achieved when the starting node, associated with the solution having the minimum length in these 10,000 TSP instances, exactly matches the node indicated by the pointer $\bm{\beta}$. Conversely, the mean solution length is $\frac{6.671}{5.752}-1\approx15.97\%$ worse than the solution length derived based on the \sysname{}'s determined pointer~$\bm{\beta}$. These results demonstrate that the solution derived by pointer $\bm{\beta}$ closely approximates the optimal decision (i.e., the starting node associated with the solution has the minimum length). Moreover, there exists a significant gap of up to $\frac{7.466}{5.748}-1\approx29.88\%$ between the solution lengths obtained using different starting nodes, underscoring the importance of selecting the appropriate one. This result further justifies our primary motivation behind the design of the starting-node pointer.


\begin{table}[!t]
  \centering
  \caption{Effectiveness comparison of the original RL strategy and our enhanced RL strategy}
  \resizebox{1\columnwidth}{!}{
    \begin{tabular}{|cccc|}
    \toprule
    \multicolumn{1}{|c}{\textbf{strategy}} & \multicolumn{1}{c}{\textbf{\tabincell{c}{GPU Memory\\usage~(MB)}}$\downarrow$} & \multicolumn{1}{c}{\textbf{\tabincell{c}{training time\\per epoch~(min)}}$\downarrow$} & \multicolumn{1}{c|}{\textbf{\tabincell{c}{average\\length}$\downarrow$}} \\
    \midrule
    original RL & 4906.95 & 12.27 & 5.780 \\
    \textbf{enhanced RL~(ours)} & \textbf{4465.12} & \textbf{9.71} & \textbf{5.752} \\
    \bottomrule
    \end{tabular}}
    \label{table6}
\end{table}%

We further compare the performance of \sysname{} utilizing our enhanced RL strategy with the original RL strategy used in conventional models \cite{Kool2019, Bresson2021, Ma2020}. As introduced in Section~\ref{training}, the original RL strategy uses two sub-modules with the identical architecture, while ours only uses one. As shown in Table~\ref{table6}, our enhanced RL strategy is more resource-efficient ($\frac{4906.95}{4465.12}-1\approx9.89\%$ lesser required GPU memory) and yields better results ($\frac{12.27}{9.71}-1\approx26.36\%$ shorter training time and $\frac{5.780}{5.752}-1\approx0.48\%$ shorter length) than the original RL strategy.

\section{Conclusion}
\label{section:Conclusion}
Our study proposes \sysname, the first NAR model trained with RL for solving TSPs. Through extensive experimentation comparing \sysname{} with SOTA models, we demonstrate that \sysname{} outperforms existing approaches in terms of solution quality, inference latency, and generalization ability. Our findings strongly suggest that \sysname{} is by far the most suitable method for solving TSPs in real-time decision-making scenarios. Moreover, we believe our approach will provide valuable insights towards solving other COPs.

While \sysname{} demonstrates superior inference speed attributed to its one-shot nature of NAR decoding, it does not present a significant advantage over the SOTA AR model \cite{Bresson2021} in terms of solution quality. To achieve a better trade-off between inference latency and solution quality, we plan to develop a semi auto-regressive variant of \sysname, which will generate multiple nodes at each decoding step by utilizing the neighboring nodes' information.



\bibliography{MyBib1}

\vfill

\end{document}